\documentclass[10pt,twocolumn,letterpaper]{article}

\usepackage[pagenumbers]{cvpr} 

\PassOptionsToPackage{dvipsnames,table}{xcolor}
\usepackage{xcolor}
\usepackage{graphicx}
\usepackage{booktabs}
\usepackage{amsmath}
\usepackage{multirow}
\usepackage{makecell}
\usepackage{csquotes}
\usepackage{wrapfig}
\usepackage{diagbox}

\usepackage[utf8]{inputenc}
\usepackage{color,soul}
\usepackage[T1]{fontenc}

\usepackage{lineno}

\definecolor{trashbin_blue}{RGB}{23,211,253}
\definecolor{hydrant_green}{RGB}{37,253,53}
\definecolor{bench_orange}{RGB}{255,195,45}

\setul{0.5ex}{0.3ex}
\setulcolor{trashbin_blue}

\usepackage{CJKutf8}

\definecolor{mygray}{RGB}{230,230,230}

\definecolor{clipcolor}{gray}{1.0}

\definecolor{openclipcolor}{gray}{1.0}

\definecolor{evaclipcolor}{gray}{1.0}

\definecolor{siglipcolor}{gray}{1.0}

\definecolor{llavacolor}{gray}{1.0}

\definecolor{sizescolor}{RGB}{194,215,255}

\definecolor{sizemcolor}{RGB}{165,202,255}

\definecolor{resscolor}{RGB}{194,255,191}

\definecolor{resmcolor}{RGB}{152,243,165}

\usepackage[most]{tcolorbox}

\definecolor{mapcolorRGB}{RGB}{101,114,156}
\colorlet{mapcolor}{mapcolorRGB}

\definecolor{langcolorRGB}{RGB}{181,133,142}
\colorlet{langcolor}{langcolorRGB}

\definecolor{colabcolorRGB}{RGB}{112,141,129}
\colorlet{colabcolor}{colabcolorRGB}

\colorlet{visioncolor}{WildStrawberry}

\newtcbox{\cvtag}{enhanced,nobeforeafter,tcbox raise base,boxrule=0.4pt,top=0mm,bottom=0mm,
  right=0mm,left=4mm,arc=2pt,boxsep=2pt,before upper={\vphantom{dlg}},
colframe=visioncolor!50!black,coltext=visioncolor!25!black,colback=visioncolor!10!white,
  overlay={\begin{tcbclipinterior}\fill[visioncolor!80] (frame.south west)
    rectangle node[text=white,font=\sffamily\bfseries\tiny,rotate=90] {CV} ([xshift=4mm]frame.north west);\end{tcbclipinterior}}}

\newtcbox{\sharetag}{enhanced,nobeforeafter,tcbox raise base,boxrule=0.4pt,top=0mm,bottom=0mm,
  right=0mm,left=4mm,arc=2pt,boxsep=2pt,before upper={\vphantom{dlg}},
colframe=langcolor!50!black,coltext=langcolor!25!black,colback=langcolor!10!white,
  overlay={\begin{tcbclipinterior}\fill[langcolor!80] (frame.south west)
    rectangle node[text=white,font=\sffamily\bfseries\tiny,rotate=90] {TRA} ([xshift=4mm]frame.north west);\end{tcbclipinterior}}}

\newtcbox{\navitag}{enhanced,nobeforeafter,tcbox raise base,boxrule=0.4pt,top=0mm,bottom=0mm,
  right=0mm,left=4mm,arc=2pt,boxsep=2pt,before upper={\vphantom{dlg}},
colframe=colabcolor!50!black,coltext=colabcolor!25!black,colback=colabcolor!10!white,
  overlay={\begin{tcbclipinterior}\fill[colabcolor!80] (frame.south west)
    rectangle node[text=white,font=\sffamily\bfseries\tiny,rotate=90] {NAV} ([xshift=4mm]frame.north west);\end{tcbclipinterior}}}

\newtcbox{\loctag}{enhanced,nobeforeafter,tcbox raise base,boxrule=0.4pt,top=0mm,bottom=0mm,
  right=0mm,left=4mm,arc=2pt,boxsep=2pt,before upper={\vphantom{dlg}},
colframe=mapcolor!50!black,coltext=mapcolor!25!black,colback=mapcolor!10!white,
    overlay={\begin{tcbclipinterior}\fill[mapcolor!80] (frame.south west)
        rectangle node[text=white,font=\sffamily\bfseries\tiny,rotate=90] {LOC} ([xshift=4mm]frame.north west);\end{tcbclipinterior}}}

\newcommand{\shareul}[1]{\setulcolor{langcolor}\ul{#1}}
\newcommand{\navul}[1]{\setulcolor{colabcolor}\ul{#1}}
\newcommand{\locul}[1]{\setulcolor{mapcolor}\ul{#1}}

\tcbuselibrary{skins}

\AtBeginEnvironment{tcolorbox}{\footnotesize}

\usepackage[noorphans,vskip=1em,leftmargin=1em]{quoting}

\definecolor{cvprblue}{rgb}{0.21,0.49,0.74}

\usepackage{graphicx}
\usepackage{subcaption}
\usepackage{makecell}
\usepackage{amsmath}
\usepackage{setspace}
\usepackage{colortbl}
\usepackage{longtable}

\usepackage{pifont}
\usepackage{multirow}
\usepackage{makecell}

\usepackage{changepage}
\usepackage{lipsum}
\usepackage[font={small}]{caption}
\usepackage{wrapfig}
\usepackage{titletoc}
\usepackage{gensymb}

\usepackage{flushend}

\definecolor{fun}{RGB}{212,226,237}
\definecolor{geom}{RGB}{219,221,206}

\definecolor{citecolor}{RGB}{65,105,225}
\usepackage[pagebackref=true,breaklinks=true,letterpaper=true,colorlinks,
citecolor=citecolor,bookmarks=false]{hyperref}

\definecolor{colorbestLA}{RGB}{249,188,187}
\definecolor{colorbestLB}{RGB}{233,181,148}
\definecolor{colorbestLC}{RGB}{230,210,160}
\definecolor{colorbestLD}{RGB}{200,205,180}
\newcommand{\colorbestLA}[0]{\cellcolor{colorbestLA}}
\newcommand{\colorbestLB}[0]{\cellcolor{colorbestLB}}
\newcommand{\colorbestLC}[0]{\cellcolor{colorbestLC}}
\newcommand{\colorbestLD}[0]{\cellcolor{colorbestLD}}

\definecolor{colorbestNA}{RGB}{249,188,187}
\definecolor{colorbestNB}{RGB}{233,181,148}
\definecolor{colorbestNC}{RGB}{230,210,160}
\definecolor{colorbestND}{RGB}{200,205,180}
\definecolor{colorbestNE}{RGB}{192,210,227}
\newcommand{\colorbestNA}[0]{\cellcolor{colorbestNA}}
\newcommand{\colorbestNB}[0]{\cellcolor{colorbestNB}}
\newcommand{\colorbestNC}[0]{\cellcolor{colorbestNC}}
\newcommand{\colorbestND}[0]{\cellcolor{colorbestND}}
\newcommand{\colorbestNE}[0]{\cellcolor{colorbestNE}}

\definecolor{colorhuman}{RGB}{233,181,148}
\newcommand{\human}[0]{\tikz[baseline=-0.5ex]\fill[colorhuman] (0,0) circle (0.9ex);}

\definecolor{colormodeA}{RGB}{230,210,160}
\newcommand{\modeA}[0]{\tikz[baseline=-0.5ex]\fill[colormodeA] (0,0) circle (0.9ex);}

\definecolor{colormodeB}{RGB}{192,210,227}
\newcommand{\modeB}[0]{\tikz[baseline=-0.5ex]\fill[colormodeB] (0,0) circle (0.9ex);}

\definecolor{colorai}{RGB}{200,205,180}
\newcommand{\ai}[0]{\tikz[baseline=-0.5ex]\fill[colorai] (0,0) circle (0.9ex);}

\DeclareRobustCommand{\legendsquare}[1]{%
  \textcolor{#1}{\rule{2ex}{2ex}}%
}

\usepackage{tikz}

\definecolor{colorlowhigh}{RGB}{176,202,196}
\newcommand{\lowhigh}[0]{\tikz[baseline=-0.5ex]\fill[colorlowhigh] (0,0) circle (0.9ex);}

\definecolor{colorhighhigh}{RGB}{52,93,125}
\newcommand{\highhigh}[0]{\tikz[baseline=-0.5ex]\fill[colorhighhigh] (0,0) circle (0.9ex);}

\definecolor{colorhighlow}{RGB}{201,205,180}
\newcommand{\highlow}[0]{\tikz[baseline=-0.5ex]\fill[colorhighlow] (0,0) circle (0.9ex);}

\definecolor{colorlowlow}{RGB}{243,243,243}
\newcommand{\lowlow}[0]{\tikz[baseline=-0.5ex]\fill[colorlowlow] (0,0) circle (0.9ex);}

\def\paperID{18} 
\def\confName{CVPR}
\def\confYear{2025}

\title{Towards Autonomous Micromobility through Scalable Urban Simulation\\\vspace{-0.2in}}

\author{
    Wayne Wu$^{1*}$
    \quad
    Honglin He$^{1*}$
    \quad
    Chaoyuan Zhang$^{2}$
    \quad
    Jack He$^{1}$\\
    Seth Z. Zhao$^{1}$
    \quad
    Ran Gong$^{1}$
    \quad
    Quanyi Li$^{1}$
    \quad
    Bolei Zhou$^{1}$ \\
    $^{1}$ University of California, Los Angeles
    \quad
    $^{2}$ University of Washington \\
    \textbf{\url{https://metadriverse.github.io/urban-sim/}}
}

\begin{document}

\twocolumn[{
            \renewcommand\twocolumn[1][]{#1}
            \maketitle
            \vspace{-0.5in}
            \begin{center}
                \centering
                \includegraphics[width=1.0\textwidth]{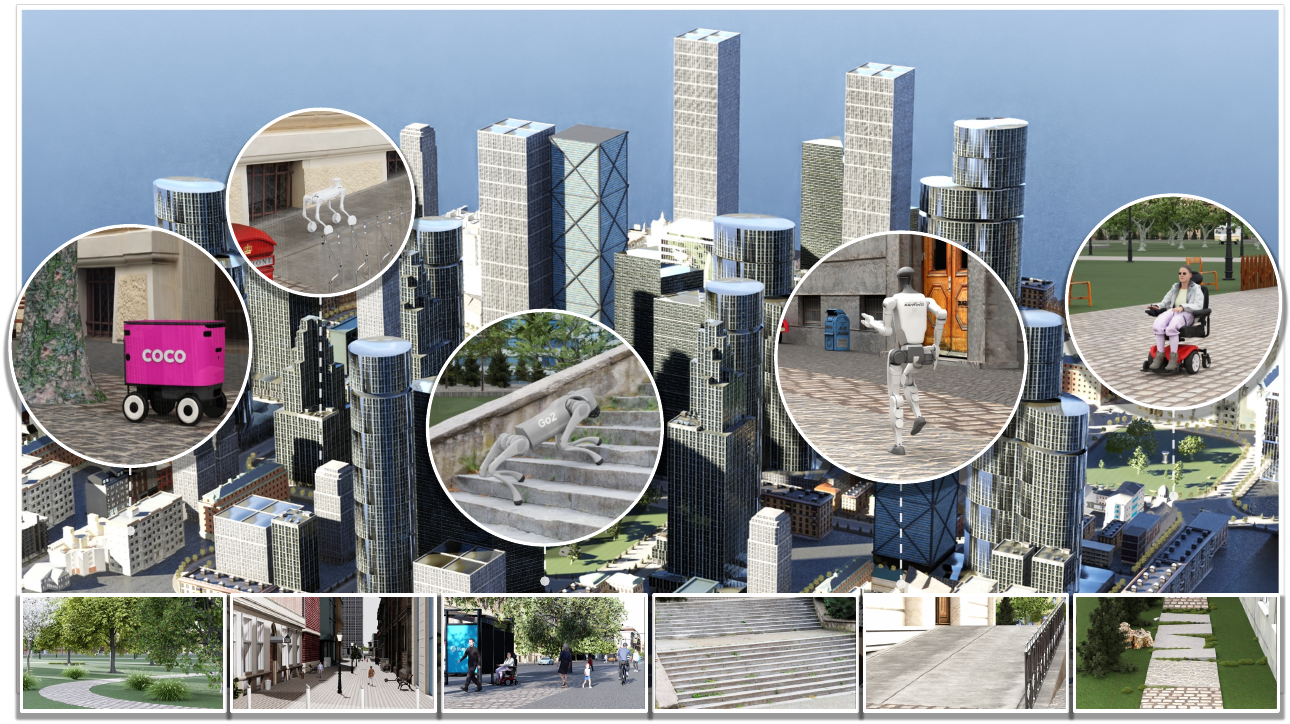}
                \vspace{-0.2in}
                \captionof{figure}{\textbf{Autonomous micromobility.} In public urban spaces, various mobile machines (circular images) are essential for short-distance travel. However, urban environments are complex and contain varied terrain and challenging situations (rectangular images). To bridge this gap, we present a scalable urban simulation solution to advance autonomous micromobility. Images are from our Urban-Tra-City data.
                }
                \label{fig:teaser}
            \end{center}
        }
        ]

\def\thefootnote{*}\footnotetext{Co-first authors.}\def\thefootnote{\arabic{footnote}}

\begin{abstract}
\vspace{-0.2in}

Micromobility, which utilizes lightweight mobile machines moving in urban public spaces, such as delivery robots and mobility scooters, emerges as a promising alternative to vehicular mobility.
Current micromobility depends mostly on human manual operation (in-person or remote control), which raises safety and efficiency concerns when navigating busy urban environments full of unpredictable obstacles and pedestrians.
Assisting humans with AI agents in maneuvering micromobility devices presents a viable solution for enhancing safety and efficiency.
In this work, we present a scalable urban simulation solution to advance autonomous micromobility.
First, we build \texttt{URBAN-SIM} -- a high-performance robot learning platform for large-scale training of embodied agents in interactive urban scenes.
\texttt{URBAN-SIM} contains three critical modules: Hierarchical Urban Generation pipeline, Interactive Dynamics Generation strategy, and Asynchronous Scene Sampling scheme, to improve the diversity, realism, and efficiency of robot learning in simulation.
Then, we propose \texttt{URBAN-BENCH} -- a suite of essential tasks and benchmarks to gauge various capabilities of the AI agents in achieving autonomous micromobility.
\texttt{URBAN-BENCH} includes eight tasks based on three core skills of the agents: Urban Locomotion, Urban Navigation, and Urban Traverse. We evaluate four robots with heterogeneous embodiments, such as the wheeled and legged robots, across these tasks. Experiments on diverse terrains and urban structures reveal each robot's strengths and limitations.

\end{abstract}
\section{Introduction}
\label{sec:introduction}

Micromobility becomes a promising urban transport way for short-distance travel~\cite{oeschger2020micromobility,abduljabbar2021role}.
It includes a range of lightweight machines that have a mass of no more than 350 \textit{kg} and operate at speeds not exceeding 45 \textit{kph}~\cite{micromobility2020report} in public spaces. These machines encompass mobile robots with different forms, such as wheeled, quadruped, wheeled-legged, and humanoid robots, and assistive mobility devices for elderly and disabled people, such as electric wheelchairs and mobility scooters. They can accommodate various users’ needs in individual travel and parcel delivery.
The appeal of micromobility lies in its provision of a flexible, sustainable, cost-effective, and on-demand transport alternative, which enhances urban accessibility~\cite{shaheen2013public,milakis2020micro} and reduces reliance on vehicles for short-distance trips~\cite{clewlow2019micro,tiwari2019micro}.

Current road designs predominantly cater to full-sized vehicles~\cite{gehl2011life}. Micromobility machines thus have to move through intricate urban public spaces, such as sidewalks, alleys, and plazas, which contain unpredictable terrains, various obstacles, and dense pedestrian traffic. Traditional micromobility machines rely on either onboard control (like wheelchairs) or teleoperation by humans (like food delivery bots~\cite{coco2024}) to navigate complex urban spaces.
However, humans and their driven mobile machines face critical \textit{safety} concerns from human fatigue and limited situational awareness. As reported by FARS~\cite{nhts2023}, over 6,000 vulnerable road users died on U.S. streets in 2018, a 14\% increase over 2015 and a 27\% increase over 2014. Humans are prone to distractions that can lead to collisions with road hazards.
On the other hand, human-driven machines have low operation \textit{efficiency}, as they require high labor costs and have limited agility. For instance, in teleoperated systems for parcel delivery~\cite{coco2024,kiwibot2024}, robots require continuous human monitoring, which limits the number of robots that can be operated simultaneously. Also, given the complexity of the urban environment, human teleoperators may find it challenging to move swiftly through a hustling street.

\textbf{Autonomous micromobility} harnesses embodied AI agents for decision-making and maneuvering, providing a viable way to improve safety and efficiency.
Existing AI solutions are mainly targeted at specific abilities of robots, such as obstacle avoidance~\cite{sorokin2022learning} and parkour~\cite{cheng2024extreme}. 
However, micromobility tasks require agents to have versatile capabilities facing various complex and challenging terrains and situations (bottom row in Figure~\ref{fig:teaser}), \ie, traversing varied terrains (stairs, slopes, and rough surfaces), moving on traversable paths in open spaces, and avoiding both static and dynamic obstacles.
Current AI solutions, focused on \textit{isolated} tasks, are thus incapable of conducting complex micromobility tasks.
Apart from that, existing robot learning and simulation platforms are insufficient for agent training on micromobility. They either have simple training scenes with \textit{no contextual environments}~\cite{makoviychuk2isaac,mittal2023orbit} or have \textit{low training performances} without environment parallelization on GPUs~\cite{dosovitskiy2017carla,li2022metadrive,wu2024metaurban}.
For example, IsaacGym~\cite{makoviychuk2isaac} has superior performance but simple environments, while CARLA~\cite{dosovitskiy2017carla} provides rich town scenes but has low end-to-end training efficiency.
However, for micromobility tasks, on the one hand, robots should learn situational awareness by interacting with large-scale scene contexts, such as urban facilities and pedestrians; on the other hand, robots need a high-performance training platform to sample diverse scenes to obtain strong generalizability. Yet, ``large-scale training'' with abundant diverse scenes and ``high-performance training'' are contradictory in the existing robot learning platforms. Current platforms can not balance these two demands and thus lack sufficient support for autonomous micromobility tasks.

In this work, we present a scalable urban simulation solution to advance autonomous micromobility. This solution consists of two critical components: a robot learning platform \textbf{\texttt{URBAN-SIM}}, and a suite of tasks and benchmarks \textbf{\texttt{URBAN-BENCH}}.
It forges a path to autonomous micromobility by enabling large-scale training and evaluation of varied embodied AI agents in complex urban environments. 

First, we propose \textbf{\texttt{URBAN-SIM}} -- a high-performance robot learning platform for autonomous micromobility. 
It can automatically construct infinite \textit{diverse} and \textit{realistic} interactive urban scenes for large-scale robot learning while providing more than 1,800 \textit{fps} high \textit{training performance} with large-scale parallelization in a single Nvidia L40S GPU.
\texttt{URBAN-SIM} has three key designs:
1) The \textit{Hierarchical Urban Generation} pipeline, which can construct an infinite number of static urban scenes, from street block to ground division to building and infrastructure placements to terrain generation. This pipeline remarkably enhances the \textit{diversity} of training environments.
2) The \textit{Interactive Dynamics Generation} strategy, which can provide rich dynamics of pedestrians and cyclists that are responsive to robots in real-time during training. This strategy highly improves the \textit{realism} of dynamic agents while maintaining the performance in our large-scale, distributed robot learning workflows.
3) The \textit{Asynchronous Scene Sampling} scheme, which can train robots on thousands of various urban scenes on GPUs in parallel. This scheme significantly enhances the \textit{training performance}, especially for large-scale scenes, achieving more than 26.3$\%$ relative improvement compared to synchronous approaches with the same training steps.
\texttt{URBAN-SIM} is built on top of Nvidia’s Omniverse~\cite{nvidia2024omniverse} and PhysX 5~\cite{nvidia2024physx} to provide realistic scene rendering and physics simulation.

Though the goal of autonomous micromobility is to move from point A to B in an urban environment, it requires the multifaceted capabilities of the agent. Thus, we construct \textbf{\texttt{URBAN-BENCH}} -- a suite of essential tasks and benchmarks to train and evaluate different capabilities of an agent.
We first construct a set of tasks for the agent to acquire two orthogonal skills in micromobility: \textit{Urban Locomotion} and \textit{Urban Navigation}.
For urban locomotion, an agent must learn various movement skills to tackle different ground conditions, \ie, flat surfaces, slopes, stairs, and rough terrain. We define four tasks for urban locomotion based on these ground conditions.
For urban navigation, an agent needs to develop different operational skills to manage various scenarios, \ie, unobstructed ground, static obstacles, and dynamic obstacles. We define three tasks for urban navigation based on these scene conditions.
Furthermore, real-world micromobility often requires \textit{kilometer-scale} navigation in complex urban spaces; it remains extremely challenging to tackle this problem.
Thus, we define \textit{Urban Traverse} as a new task with a substantially long time horizon, where a mobile robot needs to make tens of thousands of actions at a kilometer-scale distance.
We further introduce a human-AI shared autonomous approach to tackle the task. It is designed with a flexible architecture that ranges from full human control to complete AI management of the workflow, allowing us to explore various labor division modes between humans and AI agents in the urban traverse task. 

We construct comprehensive benchmarks across four robots with heterogeneous mechanical structures for all 8 defined tasks.
%
Experimental results demonstrate that all \texttt{URBAN-BENCH} tasks are challenging for existing solutions. By presenting well-defined challenges beyond the capabilities of current solutions, \texttt{URBAN-BENCH} can serve as \textit{a unified benchmark} that facilitates the future development of autonomous micromobility.
Furthermore, through training in complex urban environments, qualitative results indicate that agents have developed interesting and surprising skills based on their mechanical structures. For instance, humanoid robots learn to maneuver through narrow spaces by sidestepping, while wheeled robots learn to navigate around stairs by detouring.
Finally, we demonstrate our work's strong scale-up ability, which is essential for learning skills in autonomous micromobility.

\section{Related Work}
\label{sec:related_work}

\subsection{Micromobility}
Conventional mobility solutions~\cite{banister2008sustainable}, such as cars and buses, primarily operate on structured roadways, suited for \textit{medium} to \textit{long-distance} commutes. However, these systems often struggle with last-mile connectivity, where efficient transport is needed for the final leg of a journey, such as moving people from transit hubs to destinations or delivering parcels directly to recipients' doorsteps.
Micromobility~\cite{oeschger2020micromobility,abduljabbar2021role}, emerging in Europe and North America in the late 1900s~\cite{home1991assault,midgley2009shared}, offers a practical solution for \textit{short-distance} travel in urban spaces. It relies on lightweight and low-speed devices, such as electric wheelchairs and e-mobility scooters for personal transport~\cite{liyanage2019flexible}, or small robots for parcel delivery~\cite{demaio2009bike}, providing flexible, sustainable, and cost-effective alternatives to private vehicles. This approach reduces emissions~\cite{shaheen2010bikesharing}, alleviates congestion~\cite{masoud2019heuristic}, and enhances accessibility~\cite{shaheen2013public}, especially in densely populated areas.

Recently, a few AI-driven solutions~\cite{he2020dynamic,white2020urban} have been introduced in micromobility, focusing on device-sharing systems~\cite{tran2015modeling} and scene understanding~\cite{yang2020safety}, including fleet management, demand prediction, as well as road change and hazard detection. While these improve operational efficiency, they do not tackle the core challenge of enabling autonomous travel from point A to B in urban spaces. Current solutions lack the \textit{embodied intelligence} essential for real-time decision-making, which is crucial for tasks like assistive mobility and autonomous delivery.

\begin{figure*}[t!]
    \centering
    \includegraphics[width=1\linewidth]{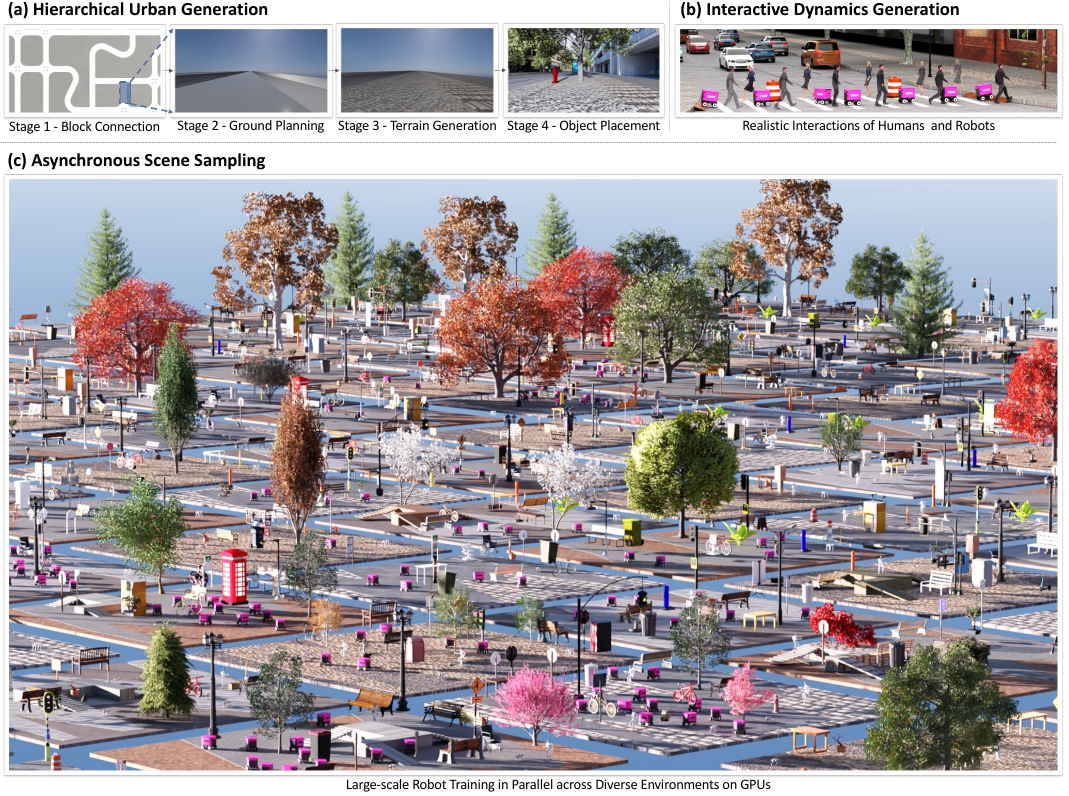}
    \caption{\textbf{\texttt{URBAN-SIM}: a robot learning platform for autonomous micromobility.}
    (a) Hierarchical Urban Generation. It generates an infinite number of \textit{diverse} scenes through four progressive stages. (b) Interactive Dynamics Generation. GPU-based generation of \textit{realistic} agent-scene and agent-agent interactions on the fly. (c) Asynchronous Scene Sampling. An asynchronous sampling scheme to enable \textit{high-efficiency} training on varied scenes with rich context information.
    }
    \label{fig:urban_sim}
\end{figure*}

\subsection{Simulation Platforms for Robot Learning}

Simulation platforms have rapidly advanced over the past decades, offering scalable training for embodied agents and robots, as well as safe evaluation before real-world deployment~\cite{li2021igibson,deitke2020robothor,deitke2022️procthor,szot2021habitat,puig2023habitat}. 
Existing platforms mainly focus on two types of environments: 1) \textit{indoor} environments~\cite{puig2018virtualhome,savva2019habitat}, such as homes and offices, and 2) \textit{driving} environments~\cite{kazemkhani2024gpudrive,krajzewicz2002sumo}, like roadways and highways.
In indoor environments, platforms like AI2-THOR~\cite{kolve2017ai2}, Habitat~\cite{savva2019habitat}, iGibson~\cite{shen2021igibson}, OmniGibson~\cite{li2024behavior}, and ThreeDWorld~\cite{gan2020threedworld} are tailored for tasks like indoor navigation, object rearrangement, and manipulation, which differ greatly from micromobility scenarios in complex urban spaces.
In driving environments, platforms like GTA V~\cite{martinez2017beyond}, CARLA~\cite{dosovitskiy2017carla}, DriverGym~\cite{kothari2021drivergym}, Nuplan~\cite{caesar2021nuplan}, and MetaDrive~\cite{li2022metadrive} support medium to long-distance driving tasks, focusing on vehicle-centric road scenarios rather than urban public spaces like sidewalks and alleys, which are crucial for micromobility tasks.

Some recent works have constructed detailed urban spaces~\cite{gao2024embodiedcity,wang2024grutopia,xie2024citydreamer,zhang2024cityx}. However, these focus mainly on algorithm evaluation~\cite{gao2024embodiedcity,wang2024grutopia} or scene generation~\cite{xie2024citydreamer,zhang2024cityx}, and lack support for interactive robot training, which requires efficient scene sampling, physical simulation, and real-time dynamics.
Recently, task-oriented robot learning platforms, such as IsaacGym~\cite{makoviychuk2isaac}, IsaacSim~\cite{nvidia2024isaacsim}, and IsaacLab~\cite{mittal2023orbit}, built on Nvidia ecosystem, have shown impressive training efficiency with high visual and physical realism. However, these platforms are mainly suited for repetitive tasks in \textit{uniform} environments, like locomotion and manipulation, and often neglect contextual scene simulation needed for complex, long-horizon micromobility tasks.

\subsection{Robot Autonomy Tasks}

Recent advances in robotics and embodied AI have significantly enhanced specific skills for robot autonomy, particularly in \textit{locomotion}~\cite{kotha2024next} and \textit{navigation}~\cite{desouza2002vision}.
In locomotion, the main goal is to enable robots to move efficiently across diverse terrains.
Considerable progress has been achieved in tasks categorized by different mechanical structures (\eg, bipedal~\cite{li2023robust}, quadrupedal~\cite{agarwal2023legged}, multilegged~\cite{chong2023multilegged}) or unique abilities (\eg, parkour~\cite{cheng2024extreme}, whole-body control~\cite{liu2024visual}, jumping~\cite{smith2023learning}).
In navigation, the focus is on guiding robots to specific destinations while avoiding obstacles. Research has proposed various tasks categorized by goals and conditions, such as point navigation~\cite{anderson2018evaluation}, object navigation~\cite{yokoyama2024hm3d}, and social navigation~\cite{tsoi2022sean}.
However, these tasks address \textit{isolated} skills and struggle to meet micromobility’s demands, which require unique and versatile abilities for complex urban environments.
A few pioneering studies have explored long-horizon outdoor navigation tasks, but they are limited to \textit{case-specific} robots~\cite{miki2022learning,lee2024learning} and scenarios~\cite{sorokin2022learning,shah2022viking}, lacking the generalizability needed for micromobility tasks.
In this work, we evaluate holistic tasks across different robots, from foundational abilities like locomotion and navigation to comprehensive tasks like traverse, which are essential for advancing autonomous micromobility in urban environments.
\section{\texttt{URBAN-SIM}: A Robot Learning Platform for Autonomous Micromobility}
\label{sec:urban_sim}

%
To support robot learning in complex urban scenes, an ideal simulation platform needs to have two important features: \textbf{large-scale} -- the platform should provide a vast array of \textit{diverse} scenes with \textit{realistic} interactions; and \textbf{high-performance} -- the platform should support \textit{high-efficiency} scene sampling for training.
In this section, we introduce \texttt{URBAN-SIM} -- a robot learning platform for autonomous micromobility, which can balance the contradiction between scale and performance. It supports infinite urban scene generation with arbitrary size and achieves high-performance training with more than 1,800 \textit{fps} sampling rate in a single GPU.
We highlight three key designs of \texttt{URBAN-SIM}: the \textbf{Hierarchical Urban Generation} pipeline (Section~\ref{sec:scene_generation}), which ensures the \textit{diversity} of static scenes on a large scale; the \textbf{Interactive Dynamics Generation} strategy (Section~\ref{sec:dynamic_generation}), which ensures the \textit{realism} of dynamics on a large scale; and the \textbf{Asynchronous Scene Sampling} scheme (Section~\ref{sec:scene_sampling}), which ensures \textit{high-efficiency} training on complex urban environments.

\subsection{Hierarchical Urban Generation}
\label{sec:scene_generation}

The \textit{diversity} of simulation environments is essential for the robustness and generalizability of robot training, especially in deep learning approaches.
Following recent advancements in procedural generation in games~\cite{short2017procedural}, we introduce a hierarchical urban generation pipeline to procedurally create complex urban scenes, from macroscale street blocks to microscale terrains, enabling \textit{infinite variations} of diverse scenes with \textit{arbitrary sizes} (from a street corner to a city).

As shown in Figure~\ref{fig:urban_sim} (a), this pipeline includes four progressive stages:
1) In block connection, street blocks (\eg, straight, curve, roundabout, diverging, merging, intersection, and T-intersection) are sampled and connected to form diverse road networks.
2) In ground planning, we divide urban public areas into functional zones (\eg, sidewalks, crosswalks, plazas, buildings, and vegetation) using randomized parameters for each area’s dimensions.
3) In terrain generation, we employ the Wave Function Collapse (WFC)~\cite{gumin2016wave} algorithm to generate typical urban terrains - flat (\eg, pathway on grass), stair (\eg, front steps), slope (\eg, assistive ramps), and rough (\eg, cracked sidewalks) - each with adjustable parameters like step height or ramp angle, providing diverse ground conditions.
4) In object placement, static objects (\eg, trees and bus stops) are placed adaptably within the functional areas according to their sizes, creating varied obstacle layouts. To ensure the coverage of objects, we have compiled a repository of over 15,000 high-quality 3D assets of urban objects.
This pipeline enables the creation of enormous static urban scenes with diverse street layouts, functional divisions, obstacles, and terrains in a breeze\footnote{Empowered by the UI of Omniverse~\cite{nvidia2024omniverse}, users can easily modify the scenes generated by our pipeline further, to cater to specific needs.}.

\subsection{Interactive Dynamics Generation}
\label{sec:dynamic_generation}

Beyond static scene diversity, the \textit{realism} of dynamic agents, \ie, vehicles, pedestrians, and other mobile machines, is crucial for simulated urban environments.
To form realistic dynamics, the environmental agents should be interactive, with both the static scenes and other dynamic agents.
A naive approach uses multi-agent path planning algorithms like ORCA~\cite{van2011reciprocal} to optimize agents' trajectories, avoiding collisions and deadlocks. However, these methods pre-compute trajectories, preventing real-time interaction with the trained agent, and run only on the CPU, causing inefficiencies when integrated with GPU-based platforms due to the frequent CPU-GPU data transfer during training.

To address these issues, we follow Waymax~\cite{gulino2024waymax} and JaxMARL~\cite{rutherford2023jaxmarl} in upgrading ORCA with JAX~\cite{bradbury2018jax} for multi-agent path planning on GPUs without any CPU bottlenecks. This method enables parallelization across multiple environments for simultaneous collision avoidance with static and dynamic objects and interaction with the trained agent.
Specifically, we first generate a 2D occupancy map labeling obstacles, roadways (for vehicles), and traversable areas (for pedestrians and mobile machines), then sample random start and end points for each agent. Using ORCA for initial trajectories, we adjust agents' positions in real-time based on proximity and relative velocity, all on GPUs. We illustrate the realistic interactions between agents and environments and other agents in Figure~\ref{fig:urban_sim} (b).
This strategy enables the creation of dynamic environments with realistic interactions on the fly in robot training.

\subsection{Asynchronous Scene Sampling}
\label{sec:scene_sampling}

So far, we can generate diverse scenes with realistic dynamics. However, the complexity of these scenes, with numerous objects and dense physical interactions, poses new challenges for the training performance, especially in learning long-horizon behaviors for robots with high degrees of freedom.
Recent robot learning platforms like IsaacGym~\cite{makoviychuk2isaac} and IsaacLab~\cite{mittal2023orbit} achieve high performance through environment parallelization on GPUs. These platforms are designed for tasks that require extensive repetitive training in \textit{uniform} environments with enormous trial and error, such as locomotion and manipulation.
In micromobility tasks, however, rather than uniform environments, robots must make decisions based on \textit{varied} environments with rich contextual information, such as ground paving, obstacles semantics, and pedestrian movements.
Thus, existing \textit{synchronous} scene sampling in \cite{makoviychuk2isaac,mittal2023orbit} will encounter huge barriers facing micromobility tasks, where \textit{the essential is not the repetitive training in uniform environments but the multi-faceted training in enormous varied environments}.

\begin{figure}[h!]
    \centering
    \includegraphics[width=1\linewidth]{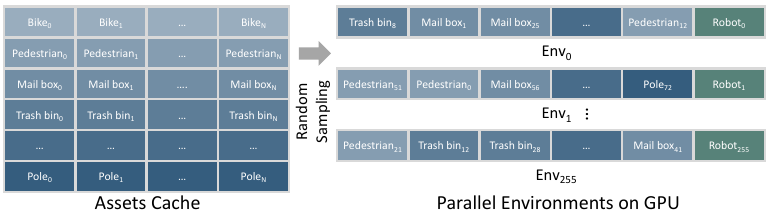}
    \vspace{-0.25in}
    \caption{\textbf{Scene sampling diagram.}
    (Left) Assets Cache that stores all assets in urban scenes. (Right) With a random sampling of assets, parallel environments can be constructed on GPU.
    }
    \vspace{-0.1in}
    \label{fig:asyn_scene_sampling}
\end{figure}

To solve this problem, we propose an \textit{asynchronous} scene sampling scheme, which can remarkably enhance training efficiency by training simultaneously on thousands of non-uniform environments with various static layouts, obstacles, dynamics, terrains, and episodes of agents.
Specifically, as illustrated in Figure~\ref{fig:asyn_scene_sampling}, all assets are initially loaded into a cache, from which environments randomly sample assets to create diverse settings simultaneously. Observations, rewards, and actions for each environment are fully vectorized on the GPU, enabling efficient parallel training of agents across multiple environments.
Figure~\ref{fig:urban_sim} (c) visualizes the parallel training on varied environments simultaneously with the asynchronous scene sampling scheme.
This approach significantly accelerates model convergence and reduces training time, essential for context-aware micromobility tasks.

\paragraph{Performance benchmarking.}
Using the asynchronous scene sampling scheme, we can enable parallelization with any number of unique environments, depending on the GPU used. On a single GPU, parallelized training can be conducted across 256 environments, achieving performance ranging from 1,800 to 2,600 \textit{fps} with RGBD sensors, depending on the specific scenario. 
Note that, due to the scalable nature of our platform, the sampling rate can be continually increased by adding more GPUs.
Please refer to the \underline{Appendix} for detailed performance benchmarks.

\section{\texttt{URBAN-BENCH}: A Suite of Essential Tasks for Autonomous Micromobility}
\label{sec:urban_bench}

%
In this section, we introduce \texttt{URBAN-BENCH}, a suite of essential tasks and benchmarks that capture high-frequency scenarios in autonomous micromobility.
Based on the data from users of micromobility, we first summarize several key \textbf{Human Needs} (Section~\ref{sec:human_needs}) as the basis of the task definition.
The real-world demands for micromobility devices mainly ask for two primary skills: 
\textbf{Urban Locomotion} (Section~\ref{sec:urban_locomotion}) — moving stably across diverse terrains, including flat, slope, stair, and rough surfaces, and
\textbf{Urban Navigation} (Section~\ref{sec:urban_navigation}) — moving efficiently in spaces with varying conditions like unobstructed pathways, static, and dynamic obstacles.
Furthermore, we define a long-horizon task, \textbf{Urban Traverse} (Section~\ref{sec:urban_traverse}), where robots must navigate urban spaces at kilometer scales.
To tackle this challenging task, we introduce a pilot approach - human-AI shared autonomy - leveraging the power of both humans and AI agents.
We will present benchmark results for these tasks in Section~\ref{sec:benchmarks}.

\begin{figure*}[h!]
    \centering
    \includegraphics[width=1\linewidth]{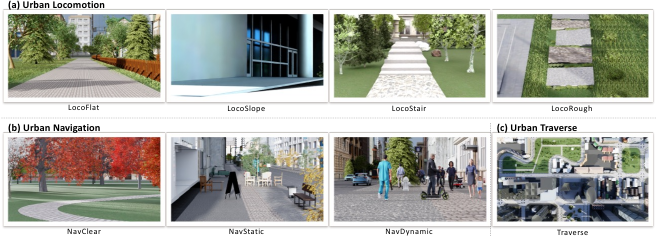}
    \caption{\textbf{\texttt{URBAN-BENCH}: a suite of essential tasks for autonomous micromobility.} Simulation environments of eight essential tasks of (a) Urban Locomotion, (b) Urban Navigation, and (c) Urban Traverse.
    }
    \label{fig:urban_bench}
\end{figure*}

\subsection{Tasks Grounded in Human Needs}
\label{sec:human_needs}

The selection of tasks in \texttt{URBAN-BENCH} is informed by urban mobility studies and infrastructure assessments, highlighting their practical importance.
U.S. Department of Transportation (DOT) reports~\cite{dot} indicate the prevalence of diverse terrains like ramps, stairs, and uneven surfaces in public spaces, so it is necessary to have various locomotion capabilities, including \textit{slope traversal}, \textit{stair climbing}, and \textit{rough terrain traversal}.
Besides, the National Household Travel Survey (NHTS)~\cite{nhts} indicates that a significant portion of urban travel involves short trips on sidewalks and plazas, where micromobility devices must navigate both unobstructed pathways and crowded zones. This underscores the need for safe and efficient \textit{clear pathway traversal}, and \textit{static} and \textit{dynamic obstacle avoidance}.
Based on these scene conditions, we define a set of essential tasks of urban locomotion and navigation.

\subsection{Urban Locomotion}
\label{sec:urban_locomotion}

In urban locomotion, the embodied AI agent controls the robot's locomotion, ensuring stable and efficient movement across various terrains such as flat surfaces, slopes, and stairs.
We define four tasks for urban locomotion (Figure~\ref{fig:urban_bench} (a)) based on different ground conditions as below:

\noindent
\loctag{\texttt{LocoFlat}} $\to$ \locul{Flat Terrain Traversal}:
Traversing stable, flat surfaces commonly found on sidewalks and pedestrian zones. This is necessary for basic mobility in city spaces designed for foot traffic.

\noindent
\loctag{\texttt{LocoSlope}} $\to$ \locul{Incline Ascent and Descent}:
Moving up and down ramps and inclined surfaces with varying slope angles. This is essential in urban areas where slopes and accessibility ramps are common.

\noindent
\loctag{\texttt{LocoStair}} $\to$ \locul{Stair Ascent and Descent}:
Ascending and descending stairs with varying heights. This is critical in urban spaces where ramps are unavailable, allowing access to multi-level areas.

\noindent
\loctag{\texttt{LocoRough}} $\to$ \locul{Uneven Terrain Traversal}:
Maintaining stability on uneven surfaces like cobblestones or damaged sidewalks. This is important for robust movement in urban environments with irregular, worn-down paths.

\subsection{Urban Navigation}
\label{sec:urban_navigation}

In urban navigation, the embodied AI agent handles local navigation, determining how the robot should move to stay within traversable areas while avoiding obstacles and pedestrians.
We define three tasks for urban navigation (Figure~\ref{fig:urban_bench} (b)) based on different scene conditions as below:

\noindent
\navitag{\texttt{NavClear}} $\to$ \navul{Traversable Area Finding}:
Moving across open, unobstructed ground, avoiding non-walkable areas like mud or bushes. This is essential for efficient navigation on open plazas and trails on lawns.

\noindent
\navitag{\texttt{NavStatic}} $\to$ \navul{Static Obstacle Avoidance}:
Navigating around stationary urban obstacles such as benches, trash bins, and signposts. This is vital for safely maneuvering in crowded city environments with fixed structures.

\noindent
\navitag{\texttt{NavDynamic}} $\to$ \navul{Dynamic Obstacle Avoidance}:
Adjusting paths to avoid moving obstacles like pedestrians and cyclists. This is crucial in urban spaces with high foot traffic, ensuring safe interactions with moving entities.

\subsection{Urban Traverse}
\label{sec:urban_traverse}

In kilometer-scale urban traverse, the embodied AI agent's goal is to reach the target point as efficiently as possible, minimizing travel time while ensuring safety in the journey.
We define the urban traverse task (Figure~\ref{fig:urban_bench} (c)) as below: 

\noindent
\sharetag{\texttt{Traverse}} $\to$ \shareul{Urban Traverse}: Moving from point A to point B with a distance of more than 1 $km$ within a complex urban environment safely and efficiently. A challenging real-world setting for micromobility.

\paragraph{Human-AI shared autonomous approach.} We propose a human-AI shared autonomous approach as a pilot study to address this task, combining AI capabilities with human interventions.
In this approach, we structure the robot control into three layers: high-level decision-making, mid-level navigation, and low-level locomotion. With the layered architecture, we decompose the complex urban traverse task into a series of subtasks, with AI managing mid-level and low-level routine tasks, and humans making high-level decisions and intervening in risky situations.
This approach allows a flexible transition between human and AI control. Humans can manage the entire process if needed, while AI can manage the entire operation using an extra rule-based/AI-based decision model to direct the dispatch of urban navigation and locomotion models.
We evaluate these control variants to study micromobility performance at the kilometer scale in Section~\ref{sec:benchmarks}.
Please refer to the \underline{Appendix} for a detailed discussion of this approach.

\section{Benchmarks}
\label{sec:benchmarks}

We benchmark four tasks in urban locomotion, three tasks in urban navigation, and one long-horizon task in urban traverse.
We describe the benchmarks below regarding the \textbf{Settings} (Section~\ref{sec:bench_settings}) of robots, data, and models, as well as the analysis of the \textbf{Results} (Section~\ref{sec:bench_results}) of benchmarks.
These benchmarks will be maintained and updated as time goes on to cover more robots, tasks, and models, as we aim to build a standard evaluation platform to facilitate research in autonomous micromobility and robot learning in urban spaces.
Please see the \underline{Appendix} for more details, including data, training parameters, evaluation metrics, \etc.

\subsection{Settings}
\label{sec:bench_settings}

\paragraph{Robots.} We evaluate four representative robots, each with distinct mechanical structures, to gain insights and demonstrate the general applicability of the proposed platform. The robots selected for this study include a wheeled robot (COCO Robotics' delivery robot), a quadruped robot (Unitree Go2), a wheeled-legged robot (Unitree B2-W), and a humanoid robot (Unitree G1)
\footnote{It is simple to import new robots in \texttt{URBAN-SIM}.}.

\paragraph{Data.} We construct 4 datasets in \texttt{URBAN-SIM}:
Urban-Nav is used for the training and testing of urban navigation; Urban-Loc is used for the training and testing of urban locomotion; Urban-Tra-Standard and Urban-Tra-City are used for the testing of urban traverse.

\paragraph{Models.} 
For the urban navigation and locomotion task, we formulate it as a Markov Decision Process (MDP)~\cite{puterman1990markov}, where the AI learns to optimize its navigation or locomotion policy using the reinforcement learning algorithm Proximal Policy Optimization (PPO)~\cite{schulman2017proximal}.
For each robot, we train and test three models for urban navigation tasks on Urban-Nav and four models for urban locomotion on Urban-Loc (except wheeled devices), which form a \textit{24-model matrix}.
%
For the urban traverse task, we construct 4 control modes, spanning from the full human to full AI:
Human -- a full human control mode;
Human-AI-Mode-1 -- a human AI shared control mode with the dispatch of foundational navigation and locomotion models;
Human-AI-Mode-2 -- a human AI shared control mode with the dispatch of foundational navigation models and a general locomotion model;
AI -- a full AI control model.

\subsection{Results}
\label{sec:bench_results}

\begin{figure*}[t!]
    \centering
    \includegraphics[width=1\linewidth]{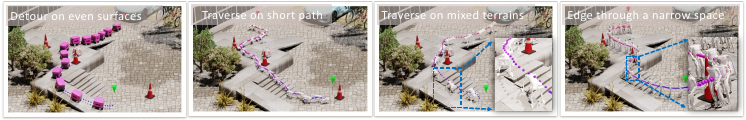}
    \vspace{-0.2in}
    \caption{\textbf{Emerging behaviors.} The results of evaluating different robots in the same environment. After training in diverse urban scenes, robots with distinct structures have developed their unique movement skills.
    }
    \vspace{-0.1in}
    \label{fig:qualitative_results}
\end{figure*}

\paragraph{Urban locomotion benchmark.}

Table~\ref{tab:locomotion_benchmark} brings the following insights:
\textit{1) Quadruped robot achieves optimal smoothness}: The quadruped robot consistently demonstrates the best Smoothness scores across all terrains, highlighting its stability and controlled movement, even on challenging surfaces like stairs and rough ground.
\textit{2) Wheeled-legged robot excels in versatility}: Leveraging its hybrid leg-wheel design, the wheeled-legged robot leads in both distance traversal ($X$-displacement and Time to Fall) and keeping Balance, enabling it to cover diverse urban terrains efficiently.
\textit{3) Humanoid robot shows stability on even surfaces}: The Humanoid robot achieves the best Balance performance on both flat and inclined ground, indicating its capability for navigation in even urban environments.

\begin{table}[h!]
\small
\caption{\textbf{Urban Locomotion benchmark.} Different colors indicate the best performance of different metrics among three robots: \legendsquare{colorbestLA} Balance; \legendsquare{colorbestLB} $X$-displacement; \legendsquare{colorbestLC} Time to Fall (TTF);  \legendsquare{colorbestLD} Smoothness.
}
\vspace{-0.1in}
\label{tab:locomotion_benchmark}
\begin{center}
\resizebox{0.47\textwidth}{!}{
    \begin{tabular}{c|c|c|c|c}
        \toprule
        \textbf{Metrics} & \loctag{LocoFlat} & \loctag{LocoSlope} & \loctag{LocoStair} & \loctag{LocoRough} \\
        \midrule
        \multicolumn{5}{c}{\includegraphics[height=0.15in]{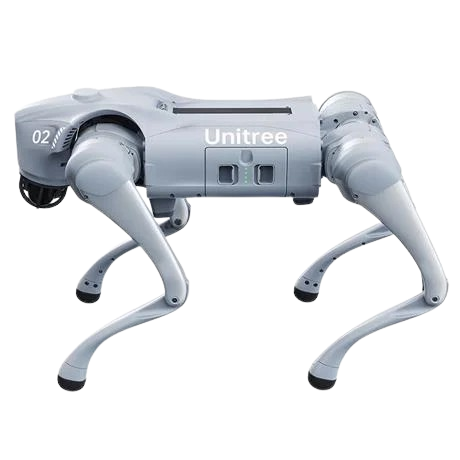} \textbf{Quadruped Robot}} \\
        \midrule
        Balance (\%) $\uparrow$ &\colorbestLA{$100.00 \pm 0.00$} &$90.56 \pm 3.13$ & \colorbestLA{$91.89 \pm 2.07$} & $72.18 \pm 4.76$ \\
        $X$-dis. (m) $\uparrow$ &$19.58 \pm 0.41$ &$4.63 \pm 0.23$ &$9.20 \pm 0.36$ &$4.88 \pm 0.14$ \\
        TTF (s) $\uparrow$ & \colorbestLC{$20.00 \pm 0.00$} &$19.50 \pm 0.44$ & \colorbestLC{$19.58 \pm 0.39$} &$18.31 \pm 0.25$ \\
        Smooth. $\downarrow$ & \colorbestLD{$7.85 \pm 0.04$} & \colorbestLD{$5.18 \pm 0.07$} & \colorbestLD{$8.11 \pm 0.12$} & \colorbestLD{$10.02 \pm 0.09$} \\
        \midrule
        \multicolumn{5}{c}{\includegraphics[height=0.15in]{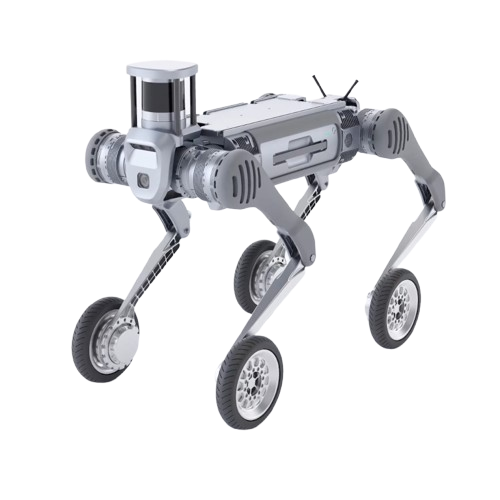} \textbf{Wheeled-Legged Robot}} \\
        \midrule
        Balance (\%) $\uparrow$ & \colorbestLA{$100.00 \pm 0.00$} &$95.57 \pm 3.31$ &$83.01 \pm 2.37$ &\colorbestLA{$85.04 \pm 2.16$} \\
        $X$-dis. (m) $\uparrow$ & \colorbestLB{$19.62 \pm 0.15$} & \colorbestLB{$12.54 \pm 0.34$} & \colorbestLB{$16.73 \pm 0.27$} & \colorbestLB{$18.24 \pm 0.22$} \\
        TTF (s) $\uparrow$ & \colorbestLC{$20.00 \pm 0.00$} & \colorbestLC{$19.95 \pm 0.02$} &$19.07 \pm 0.17$ & \colorbestLC{$19.13 \pm 0.11$} \\
        Smooth. $\downarrow$ &$210.43 \pm 0.07$ &$253.24 \pm 0.28$ &$236.52 \pm 0.18$ &$231.96 \pm 0.14$ \\
        \midrule
        \multicolumn{5}{c}{\includegraphics[height=0.15in]{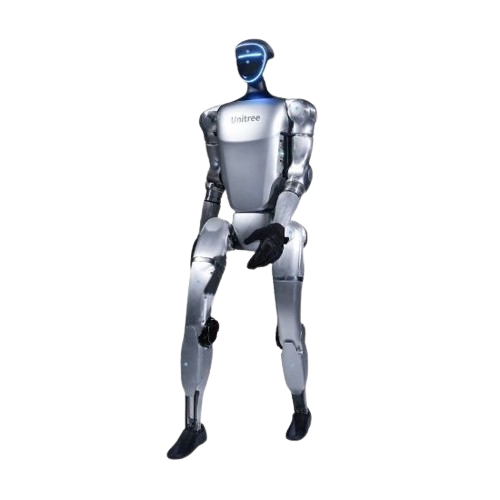} \textbf{Humanoid Robot}} \\
        \midrule
        Balance (\%) $\uparrow$ & \colorbestLA{$100.00 \pm 0.00$} & \colorbestLA{$95.67 \pm 2.24$} &$80.98 \pm 4.32$ &$82.45 \pm 3.15$ \\
        $X$-dis. (m) $\uparrow$ &$16.61 \pm 0.50$ &$7.16 \pm 0.22$ &$13.99 \pm 0.27$ &$16.28 \pm 0.31$ \\
        TTF (s) $\uparrow$ & \colorbestLC{$20.00 \pm 0.00$} & $19.91 \pm 0.03$ &$19.03 \pm 0.36$ &$19.02 \pm 0.33$ \\
        Smooth. $\downarrow$ &$40.94 \pm 0.15$ &$57.69 \pm 0.31$ &$42.36 \pm 0.19$ &$53.67 \pm 0.24$ \\
        \bottomrule
    \end{tabular}
}
\vspace{-0.15in}
\end{center}
\end{table}

\paragraph{Urban navigation benchmark.}

Table~\ref{tab:navigation_benchmark} brings the following insights.
\textit{1) Wheeled robot excels in clear pathway navigation}: The wheeled robot achieves the highest Success Rate (97.60\%) and Route Completion (98.61\%) in the \texttt{NavClear} task, highlighting its suitability for open, predictable urban environments.
\textit{2) Quadruped robot leads in safety metrics}: The quadruped robot outperforms others in tasks with obstacles, achieving the lowest Collision rates (0.08 in \texttt{NavSta} and 0.13 in \texttt{NavDyn}) and the highest percentage On Walkable Regions. This demonstrates its stability in complex, obstacle-rich environments.
\textit{3) Humanoid robot performs best in complex scenarios}: The humanoid robot shows the highest Success Rates and Route Completion in tasks with static and dynamic obstacles, indicating its flexibility in navigating crowded urban spaces.

\begin{table}[h!]
\small
\caption{\textbf{Urban navigation benchmark.} Different colors indicate the best performance of different metrics among four robots: \legendsquare{colorbestNA} Success Rate; \legendsquare{colorbestNB} Route Completion; \legendsquare{colorbestNC} On Walkable Region;  \legendsquare{colorbestND} SPL; \legendsquare{colorbestNE} Collision.
}
\vspace{-0.1in}
\label{tab:navigation_benchmark}
\begin{center}
\resizebox{0.47\textwidth}{!}{
    \begin{tabular}{c|c|c|c}
        \toprule
        \textbf{Metrics} & \navitag{\texttt{NavClear}} & \navitag{\texttt{NavStatic}} & \navitag{\texttt{NavDynamic}} \\
        \midrule
        \multicolumn{4}{c}{\includegraphics[height=0.15in]{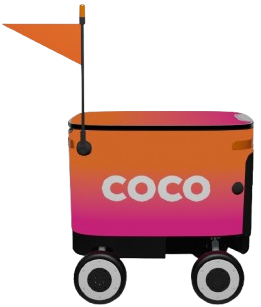} \textbf{Wheeled Robot}} \\
        \midrule
        Success Rate (\%) $\uparrow$ & \colorbestNA{$97.60 \pm 0.92$} &$51.95 \pm 2.63$ &$48.82 \pm 3.26$ \\
        Route Completion (\%) $\uparrow$ & \colorbestNB{$98.61 \pm 1.28$} &$53.11 \pm 2.92$ &$50.04 \pm 3.02$ \\
        On Walkable Region (\%) $\uparrow$ &$74.38 \pm 0.99$ &$81.88 \pm 1.00$ &$84.82 \pm 1.49$ \\
        SPL $\uparrow$ & \colorbestND{$0.48 \pm 0.05$} &$0.24 \pm 0.04$ &$0.23 \pm 0.01$ \\
        Collision $\downarrow$ &- &$0.31 \pm 0.09$ &$0.35 \pm 0.04$ \\
        \midrule
        \multicolumn{4}{c}{\includegraphics[height=0.15in]{figures/quadruped_robot.png} \textbf{Quadruped Robot}} \\
        \midrule
        Success Rate (\%) $\uparrow$ &$90.29 \pm 3.25$ &$76.13 \pm 3.07$ &$77.14 \pm 2.57$ \\
        Route Completion (\%) $\uparrow$ &$94.28 \pm 2.16$ &$77.47 \pm 2.99$ &$77.63 \pm 2.12$ \\
        On Walkable Region (\%) $\uparrow$ & \colorbestNC{$93.96 \pm 3.38$} &$85.81 \pm 1.67$ & \colorbestNC{$88.20 \pm 2.17$} \\
        SPL $\uparrow$ &$0.37 \pm 0.05$ &$0.36 \pm 0.04$ &$0.36 \pm 0.05$ \\
        Collision $\downarrow$ &- & \colorbestNE{$0.08 \pm 0.02$} & \colorbestNE{$0.13 \pm 0.02$} \\
        \midrule
        \multicolumn{4}{c}{\includegraphics[height=0.15in]{figures/wheeled_legged_robot.png} \textbf{Wheeled-Legged Robot}} \\
        \midrule
        Success Rate (\%) $\uparrow$ &$79.94 \pm 3.06$ &$42.97 \pm 4.14$ &$31.06\pm 3.77$ \\
        Route Completion (\%) $\uparrow$ &$80.44 \pm 2.97$ &$44.33 \pm 3.74$ &$33.95\pm 3.21$ \\
        On Walkable Region (\%) $\uparrow$ &$67.93 \pm 0.85$ &$62.17 \pm 2.95$ &$63.29\pm 2.71$ \\
        SPL $\uparrow$ &$0.37 \pm 0.03$ &$0.19 \pm 0.02$ &$0.14\pm 0.02$ \\
        Collision $\downarrow$ &- &$0.15 \pm 0.04$ &$0.19\pm 0.02$ \\
        \midrule
        \multicolumn{4}{c}{\includegraphics[height=0.15in]{figures/humanoid_robot.png} \textbf{Humanoid Robot}} \\
        \midrule
        Success Rate (\%) $\uparrow$ &$80.47 \pm 2.29$ & \colorbestNA{$77.86 \pm 3.54$} & \colorbestNA{$79.23 \pm 2.71$} \\
        Route Completion (\%) $\uparrow$ &$80.92 \pm 1.36$ &\colorbestNB{$79.72 \pm 2.76$} & \colorbestNB{$80.26 \pm 2.92$} \\
        On Walkable Region (\%) $\uparrow$ &$65.86 \pm 1.56$ & \colorbestNC{$86.89 \pm 1.73$} &$65.85 \pm 1.94$ \\
        SPL $\uparrow$ &$0.37 \pm 0.01$ & \colorbestND{$0.37 \pm 0.03$} & \colorbestND{$0.38 \pm 0.03$} \\
        Collision $\downarrow$ &- &$0.13 \pm 0.03$ &$0.15 \pm 0.04$ \\
        \bottomrule
    \end{tabular}
}
\vspace{-0.15in}
\end{center}
\end{table}

\paragraph{Urban traverse benchmark.}

\begin{figure}[h!]
    \centering
    \includegraphics[width=1\linewidth]{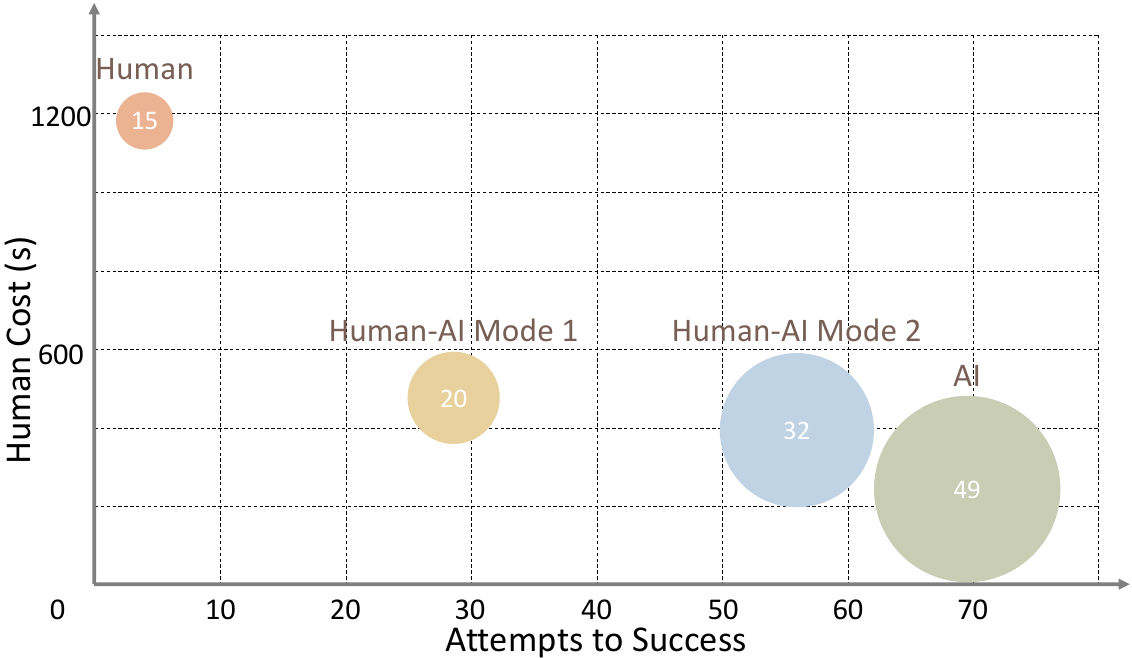}
    \caption{\textbf{Comparison of different control modes in urban traverse.} X-axis: Attempts to Success -- the number of failures before reaching the goal points (completion ability). Y-axis: Human Cost -- time of human takeover of the control (labor cost). Size of circle: Collision Times to obstacles and pedestrians (safety property). {\protect\human \protect\modeA \protect\modeB \protect\ai} indicate four control modes.
    }
    \label{fig:urban_traverse}
\end{figure}

We evaluate a quadruped robot on a kilometer-scale urban traverse task using the Urban-Tra-Standard dataset with three control modes.
As shown in Figure~\ref{fig:urban_traverse}, the AI mode achieves the lowest human intervention but exhibits the poorest completeness and safety.
Conversely, the Human mode achieves the highest completeness and safety but at a significantly higher labor cost.
The two human-AI shared autonomy modes balance completeness and cost while maintaining moderate safety.
Future research in urban traverse should aim to move the dot closer to the origin with minimal dot size, indicating optimized completeness, cost, and safety.
Please refer to the \underline{Appendix} for the complete benchmark of urban traverse.

\paragraph{Emerging robot behaviors.}

Through large-scale training in diverse urban environments, different robots obtain movement skills that exploit their unique mechanical structures, as shown in Figure~\ref{fig:qualitative_results}: quadruped robots, known to be proficient at stair climbing, can traverse challenging terrain directly to reach the goal; wheeled robots prefer detouring over even surfaces to reduce the risk of getting stuck, despite the longer path; Wheeled-legged robots benefit from their hybrid design and show the ability to partially descend on slopes and stairs simultaneously; The humanoid robot, with greater degrees of freedom, can sidestep through narrow spaces efficiently.
\section{Evaluation of Scalability}
\label{sec:evaluations}

We try to address a fundamental question underlying the strengths demonstrated in this work:
\textit{
How does the scalability of our urban simulation contribute to autonomous micromobility?
}

The proposed asynchronous scene sampling scheme in \texttt{URBAN-SIM} enables high-performance, large-scale robot training in diverse urban environments with realistic interactions.
We compare it to synchronous sampling, as used in IsaacLab~\cite{mittal2023orbit}, where all scenes in a batch are identical. In our asynchronous approach, however, each scene in a batch is unique.
Furthermore, to assess the impact of large-scale training, we vary the number of training scenes from 1 to 1,024 and observe performance changes.
All experiments are conducted using the NavStatic task.

\begin{figure}[h!]
    \centering
    \includegraphics[width=1\linewidth]{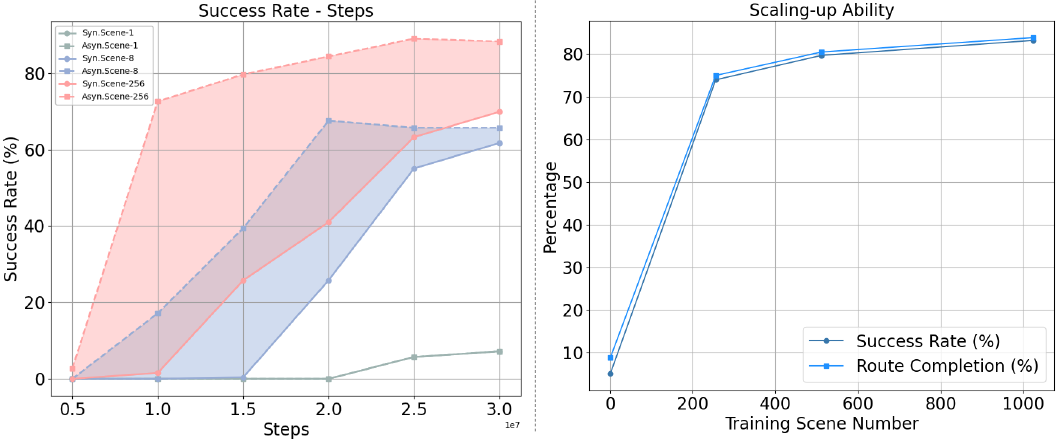}
    \caption{\textbf{Effectiveness of scalable urban simulation.} (Left) Comparison between synchronous and synchronous scene sampling. X-axis: training steps; Y-axis: Success Rate. Different colors indicate training scene numbers -- 1, 8, or 256. (Right) Scaling-up ability. X-axis: training scene number; Y-axis: Success Rate and Route Completion.
    }
    \vspace{-0.2in}
    \label{fig:evaluation_scalable_simulation}
\end{figure}

As shown in Figure~\ref{fig:evaluation_scalable_simulation} (Left), asynchronous sampling performs the same as synchronous sampling with only one scene. However, as unique training scenes increase from 8 to 256, a substantial performance gap (the colored areas) emerges, showing the strong scalability of our platform for diverse scene training.
Further, as seen in Figure~\ref{fig:evaluation_scalable_simulation} (Right), the performance remarkably improves as the number of training scenes increases from 1 to 1,024, rising from 5.1$\%$ to 83.2$\%$ (Success Rate). The result highlights the importance of large-scale training on a greater variety of scenes.

\section{Conclusion}
\label{sec:conclusion}

We introduce a scalable urban simulation solution to advance research in autonomous micromobility. This solution comprises a high-performance robot learning platform, \texttt{URBAN-SIM}, and a suite of essential tasks and benchmarks, \texttt{URBAN-BENCH}. Through experiments, we evaluate various capabilities of AI agents across different tasks and demonstrate the platform’s scalability for large-scale training in urban environments.
Looking ahead, we plan to support real-world deployments of models trained on our platform. Our strategy includes building a sim-to-real pipeline based on ROS2 and enabling an integrated workflow for model training, evaluation, and deployment.

\paragraph{Acknowledgements.}
The project was supported by the NSF grants CNS-2235012 and IIS-2339769, and ONR grant N000142512166. We extend our gratitude for the excellent assets, including 3D objects from Objaverse-XL, 3D humans from SynBody, and robots from IsaacLab.

{
    \small
    \bibliographystyle{ieeenat_fullname}
    \bibliography{camera_ready}

\begin{thebibliography}{88}
\providecommand{\natexlab}[1]{#1}
\providecommand{\url}[1]{\texttt{#1}}
\expandafter\ifx\csname urlstyle\endcsname\relax
  \providecommand{\doi}[1]{doi: #1}\else
  \providecommand{\doi}{doi: \begingroup \urlstyle{rm}\Url}\fi

\bibitem[Ren()]{Renderpeople}
Renderpeople.
\newblock \url{https://renderpeople.com/}.
\newblock Accessed: 2024-11-19.

\bibitem[Uni()]{UnityAssetStore}
Unity asset store.
\newblock \url{https://assetstore.unity.com/}.
\newblock Accessed: 2024-11.

\bibitem[Unr()]{UnrealMarketplace}
Unreal engine marketplace.
\newblock \url{https://www.unrealengine.com/marketplace/}.
\newblock Accessed: 2024-11.

\bibitem[coc()]{coco2024}
Coco robotics.
\newblock \url{https://www.cocodelivery.com/}.

\bibitem[kiw()]{kiwibot2024}
Kiwibot.
\newblock \url{https://www.kiwibot.com/}.

\bibitem[Abduljabbar et~al.(2021)Abduljabbar, Liyanage, and Dia]{abduljabbar2021role}
Rusul~L Abduljabbar, Sohani Liyanage, and Hussein Dia.
\newblock The role of micro-mobility in shaping sustainable cities: A systematic literature review.
\newblock \emph{Transportation research part D: transport and environment}, 2021.

\bibitem[Achiam et~al.(2023)Achiam, Adler, Agarwal, Ahmad, Akkaya, Aleman, Almeida, Altenschmidt, Altman, Anadkat, et~al.]{achiam2023gpt}
Josh Achiam, Steven Adler, Sandhini Agarwal, Lama Ahmad, Ilge Akkaya, Florencia~Leoni Aleman, Diogo Almeida, Janko Altenschmidt, Sam Altman, Shyamal Anadkat, et~al.
\newblock Gpt-4 technical report.
\newblock \emph{arXiv preprint arXiv:2303.08774}, 2023.

\bibitem[Agarwal et~al.(2023)Agarwal, Kumar, Malik, and Pathak]{agarwal2023legged}
Ananye Agarwal, Ashish Kumar, Jitendra Malik, and Deepak Pathak.
\newblock Legged locomotion in challenging terrains using egocentric vision.
\newblock In \emph{CoRL}, 2023.

\bibitem[Anderson et~al.(2018)Anderson, Chang, Chaplot, Dosovitskiy, Gupta, Koltun, Kosecka, Malik, Mottaghi, Savva, and Zamir]{anderson2018evaluation}
Peter Anderson, Angel~X. Chang, Devendra~Singh Chaplot, Alexey Dosovitskiy, Saurabh Gupta, Vladlen Koltun, Jana Kosecka, Jitendra Malik, Roozbeh Mottaghi, Manolis Savva, and Amir~R. Zamir.
\newblock On evaluation of embodied navigation agents.
\newblock \emph{arXiv preprint arXiv:1807.06757}, 2018.

\bibitem[Banister(2008)]{banister2008sustainable}
David Banister.
\newblock The sustainable mobility paradigm.
\newblock \emph{Transport policy}, 2008.

\bibitem[Bradbury et~al.(2018)Bradbury, Frostig, Hawkins, Johnson, Leary, Maclaurin, Necula, Paszke, VanderPlas, Wanderman-Milne, et~al.]{bradbury2018jax}
James Bradbury, Roy Frostig, Peter Hawkins, Matthew~James Johnson, Chris Leary, Dougal Maclaurin, George Necula, Adam Paszke, Jake VanderPlas, Skye Wanderman-Milne, et~al.
\newblock Jax: composable transformations of python+ numpy programs.
\newblock 2018.

\bibitem[Caesar et~al.(2021)Caesar, Kabzan, Tan, Fong, Wolff, Lang, Fletcher, Beijbom, and Omari]{caesar2021nuplan}
Holger Caesar, Juraj Kabzan, Kok~Seang Tan, Whye~Kit Fong, Eric Wolff, Alex Lang, Luke Fletcher, Oscar Beijbom, and Sammy Omari.
\newblock nuplan: A closed-loop ml-based planning benchmark for autonomous vehicles.
\newblock \emph{arXiv preprint arXiv:2106.11810}, 2021.

\bibitem[Cheng et~al.(2024)Cheng, Shi, Agarwal, and Pathak]{cheng2024extreme}
Xuxin Cheng, Kexin Shi, Ananye Agarwal, and Deepak Pathak.
\newblock Extreme parkour with legged robots.
\newblock In \emph{ICRA}, 2024.

\bibitem[Chong et~al.(2023)Chong, He, Soto, Wang, Irvine, Blekherman, and Goldman]{chong2023multilegged}
Baxi Chong, Juntao He, Daniel Soto, Tianyu Wang, Daniel Irvine, Grigoriy Blekherman, and Daniel~I Goldman.
\newblock Multilegged matter transport: A framework for locomotion on noisy landscapes.
\newblock \emph{Science}, 2023.

\bibitem[Clewlow(2019)]{clewlow2019micro}
Regina~R Clewlow.
\newblock The micro-mobility revolution: the introduction and adoption of electric scooters in the united states.
\newblock Technical report, 2019.

\bibitem[Deitke et~al.(2020)Deitke, Han, Herrasti, Kembhavi, Kolve, Mottaghi, Salvador, Schwenk, VanderBilt, Wallingford, Weihs, Yatskar, and Farhadi]{deitke2020robothor}
Matt Deitke, Winson Han, Alvaro Herrasti, Aniruddha Kembhavi, Eric Kolve, Roozbeh Mottaghi, Jordi Salvador, Dustin Schwenk, Eli VanderBilt, Matthew Wallingford, Luca Weihs, Mark Yatskar, and Ali Farhadi.
\newblock Robothor: An open simulation-to-real embodied {AI} platform.
\newblock In \emph{CVPR}, 2020.

\bibitem[Deitke et~al.(2022)Deitke, VanderBilt, Herrasti, Weihs, Ehsani, Salvador, Han, Kolve, Kembhavi, and Mottaghi]{deitke2022️procthor}
Matt Deitke, Eli VanderBilt, Alvaro Herrasti, Luca Weihs, Kiana Ehsani, Jordi Salvador, Winson Han, Eric Kolve, Aniruddha Kembhavi, and Roozbeh Mottaghi.
\newblock Procthor: Large-scale embodied ai using procedural generation.
\newblock \emph{NeuIPS}, 2022.

\bibitem[Deitke et~al.(2024)Deitke, Liu, Wallingford, Ngo, Michel, Kusupati, Fan, Laforte, Voleti, Gadre, et~al.]{deitke2024objaverse}
Matt Deitke, Ruoshi Liu, Matthew Wallingford, Huong Ngo, Oscar Michel, Aditya Kusupati, Alan Fan, Christian Laforte, Vikram Voleti, Samir~Yitzhak Gadre, et~al.
\newblock Objaverse-xl: A universe of 10m+ 3d objects.
\newblock \emph{NeuIPS}, 2024.

\bibitem[DeMaio(2009)]{demaio2009bike}
Paul DeMaio.
\newblock Bike-sharing: History, impacts, models of provision, and future.
\newblock \emph{Journal of public transportation}, 2009.

\bibitem[DeSouza and Kak(2002)]{desouza2002vision}
Guilherme~N DeSouza and Avinash~C Kak.
\newblock Vision for mobile robot navigation: A survey.
\newblock \emph{TPAMI}, 2002.

\bibitem[Dosovitskiy et~al.(2017)Dosovitskiy, Ros, Codevilla, Lopez, and Koltun]{dosovitskiy2017carla}
Alexey Dosovitskiy, German Ros, Felipe Codevilla, Antonio Lopez, and Vladlen Koltun.
\newblock Carla: An open urban driving simulator.
\newblock In \emph{CoRL}, 2017.

\bibitem[Gan et~al.(2021)Gan, Schwartz, Alter, Mrowca, Schrimpf, Traer, Freitas, Kubilius, Bhandwaldar, Haber, Sano, Kim, Wang, Lingelbach, Curtis, Feigelis, Bear, Gutfreund, Cox, Torralba, DiCarlo, Tenenbaum, McDermott, and Yamins]{gan2020threedworld}
Chuang Gan, Jeremy Schwartz, Seth Alter, Damian Mrowca, Martin Schrimpf, James Traer, Julian~De Freitas, Jonas Kubilius, Abhishek Bhandwaldar, Nick Haber, Megumi Sano, Kuno Kim, Elias Wang, Michael Lingelbach, Aidan Curtis, Kevin~T. Feigelis, Daniel Bear, Dan Gutfreund, David~D. Cox, Antonio Torralba, James~J. DiCarlo, Josh Tenenbaum, Josh~H. McDermott, and Dan Yamins.
\newblock Threedworld: {A} platform for interactive multi-modal physical simulation.
\newblock In \emph{NeurIPS Datasets and Benchmarks}, 2021.

\bibitem[Gao et~al.(2024)Gao, Zhao, Zhang, Mao, Zhang, Zheng, Man, Fang, Zhou, Cui, Chen, and Li]{gao2024embodiedcity}
Chen Gao, Baining Zhao, Weichen Zhang, Jinzhu Mao, Jun Zhang, Zhiheng Zheng, Fanhang Man, Jianjie Fang, Zile Zhou, Jinqiang Cui, Xinlei Chen, and Yong Li.
\newblock Embodiedcity: A benchmark platform for embodied agent in real-world city environment.
\newblock \emph{arXiv preprint arXiv:2410.09604}, 2024.

\bibitem[Gebru et~al.(2021)Gebru, Morgenstern, Vecchione, Vaughan, Wallach, Daum{\'e}, and Crawford]{Gebru2021DatasheetsFD}
Timnit Gebru, Jamie~H. Morgenstern, Briana Vecchione, Jennifer~Wortman Vaughan, Hanna~M. Wallach, Hal Daum{\'e}, and Kate Crawford.
\newblock Datasheets for datasets.
\newblock \emph{Comm. of the ACM}, 2021.

\bibitem[Gehl(2011)]{gehl2011life}
Jan Gehl.
\newblock Life between buildings.
\newblock 2011.

\bibitem[Gulino et~al.(2024)Gulino, Fu, Luo, Tucker, Bronstein, Lu, Harb, Pan, Wang, Chen, et~al.]{gulino2024waymax}
Cole Gulino, Justin Fu, Wenjie Luo, George Tucker, Eli Bronstein, Yiren Lu, Jean Harb, Xinlei Pan, Yan Wang, Xiangyu Chen, et~al.
\newblock Waymax: An accelerated, data-driven simulator for large-scale autonomous driving research.
\newblock \emph{NeurIPS}, 2024.

\bibitem[Gumin(2016)]{gumin2016wave}
Maxim Gumin.
\newblock Wave function collapse algorithm.
\newblock \url{https://github.com/mxgmn/}, 2016.

\bibitem[He and Shin(2020)]{he2020dynamic}
Suining He and Kang~G Shin.
\newblock Dynamic flow distribution prediction for urban dockless e-scooter sharing reconfiguration.
\newblock In \emph{Proceedings of the web conference}, 2020.

\bibitem[Home(1991)]{home1991assault}
Stewart Home.
\newblock The assault on culture: utopian currents from lettrisme to class war.
\newblock 1991.

\bibitem[Kazemkhani et~al.(2024)Kazemkhani, Pandya, Cornelisse, Shacklett, and Vinitsky]{kazemkhani2024gpudrive}
Saman Kazemkhani, Aarav Pandya, Daphne Cornelisse, Brennan Shacklett, and Eugene Vinitsky.
\newblock Gpudrive: Data-driven, multi-agent driving simulation at 1 million fps.
\newblock \emph{arXiv preprint arXiv:2408.01584}, 2024.

\bibitem[Kolve et~al.(2017)Kolve, Mottaghi, Han, VanderBilt, Weihs, Herrasti, Deitke, Ehsani, Gordon, Zhu, Aniruddha, Abhinav, and Ali]{kolve2017ai2}
Eric Kolve, Roozbeh Mottaghi, Winson Han, Eli VanderBilt, Luca Weihs, Alvaro Herrasti, Matt Deitke, Kiana Ehsani, Daniel Gordon, Yuke Zhu, Kembhavi Aniruddha, Gupta Abhinav, and Farhadi Ali.
\newblock Ai2-thor: An interactive 3d environment for visual ai.
\newblock \emph{arXiv preprint arXiv:1712.05474}, 2017.

\bibitem[Kotha et~al.(2024)Kotha, Akter, Abhi, Das, Islam, Ali, Ahamed, Islam, Sarker, Badal, et~al.]{kotha2024next}
Swapnil~Saha Kotha, Nipa Akter, Sarafat~Hussain Abhi, Sajal~Kumar Das, Md~Robiul Islam, Md~Firoj Ali, Md~Hafiz Ahamed, Md~Manirul Islam, Subrata~Kumar Sarker, Md~Faisal~Rahman Badal, et~al.
\newblock Next generation legged robot locomotion: A review on control techniques.
\newblock \emph{Heliyon}, 2024.

\bibitem[Kothari et~al.(2021)Kothari, Perone, Bergamini, Alahi, and Ondruska]{kothari2021drivergym}
Parth Kothari, Christian Perone, Luca Bergamini, Alexandre Alahi, and Peter Ondruska.
\newblock Drivergym: Democratising reinforcement learning for autonomous driving.
\newblock \emph{arXiv preprint arXiv:2111.06889}, 2021.

\bibitem[Krajzewicz et~al.(2002)Krajzewicz, Hertkorn, R{\"o}ssel, and Wagner]{krajzewicz2002sumo}
Daniel Krajzewicz, Georg Hertkorn, Christian R{\"o}ssel, and Peter Wagner.
\newblock Sumo (simulation of urban mobility)-an open-source traffic simulation.
\newblock In \emph{MESM}, 2002.

\bibitem[Lee et~al.(2024)Lee, Bjelonic, Reske, Wellhausen, Miki, and Hutter]{lee2024learning}
Joonho Lee, Marko Bjelonic, Alexander Reske, Lorenz Wellhausen, Takahiro Miki, and Marco Hutter.
\newblock Learning robust autonomous navigation and locomotion for wheeled-legged robots.
\newblock \emph{Science Robotics}, 2024.

\bibitem[Li et~al.(2021)Li, Xia, Mart{\'{\i}}n{-}Mart{\'{\i}}n, Lingelbach, Srivastava, Shen, Vainio, Gokmen, Dharan, Jain, Kurenkov, Liu, Gweon, Wu, Fei{-}Fei, and Savarese]{li2021igibson}
Chengshu Li, Fei Xia, Roberto Mart{\'{\i}}n{-}Mart{\'{\i}}n, Michael Lingelbach, Sanjana Srivastava, Bokui Shen, Kent~Elliott Vainio, Cem Gokmen, Gokul Dharan, Tanish Jain, Andrey Kurenkov, C.~Karen Liu, Hyowon Gweon, Jiajun Wu, Li Fei{-}Fei, and Silvio Savarese.
\newblock igibson 2.0: Object-centric simulation for robot learning of everyday household tasks.
\newblock In \emph{CoRL}, 2021.

\bibitem[Li et~al.(2024)Li, Zhang, Wong, Gokmen, Srivastava, Mart{\'{\i}}n{-}Mart{\'{\i}}n, Wang, Levine, Ai, Martinez, Yin, Lingelbach, Hwang, Hiranaka, Garlanka, Aydin, Lee, Sun, Anvari, Sharma, Bansal, Hunter, Kim, Lou, Matthews, Villa{-}Renteria, Tang, Tang, Xia, Li, Savarese, Gweon, Liu, Wu, and Fei{-}Fei]{li2024behavior}
Chengshu Li, Ruohan Zhang, Josiah Wong, Cem Gokmen, Sanjana Srivastava, Roberto Mart{\'{\i}}n{-}Mart{\'{\i}}n, Chen Wang, Gabrael Levine, Wensi Ai, Benjamin Martinez, Hang Yin, Michael Lingelbach, Minjune Hwang, Ayano Hiranaka, Sujay Garlanka, Arman Aydin, Sharon Lee, Jiankai Sun, Mona Anvari, Manasi Sharma, Dhruva Bansal, Samuel Hunter, Kyu{-}Young Kim, Alan Lou, Caleb~R. Matthews, Ivan Villa{-}Renteria, Jerry~Huayang Tang, Claire Tang, Fei Xia, Yunzhu Li, Silvio Savarese, Hyowon Gweon, C.~Karen Liu, Jiajun Wu, and Li Fei{-}Fei.
\newblock Behavior-1k: A human-centered, embodied ai benchmark with 1, 000 everyday activities and realistic simulation.
\newblock \emph{CoRL}, 2024.

\bibitem[Li et~al.(2022)Li, Peng, Feng, Zhang, Xue, and Zhou]{li2022metadrive}
Quanyi Li, Zhenghao Peng, Lan Feng, Qihang Zhang, Zhenghai Xue, and Bolei Zhou.
\newblock Metadrive: Composing diverse driving scenarios for generalizable reinforcement learning.
\newblock \emph{TPAMI}, 2022.

\bibitem[Li et~al.(2023)Li, Peng, Abbeel, Levine, Berseth, and Sreenath]{li2023robust}
Zhongyu Li, Xue~Bin Peng, Pieter Abbeel, Sergey Levine, Glen Berseth, and Koushil Sreenath.
\newblock Robust and versatile bipedal jumping control through reinforcement learning.
\newblock In \emph{RSS}, 2023.

\bibitem[Liu et~al.(2023)Liu, Li, Wu, and Lee]{liu2023visual}
Haotian Liu, Chunyuan Li, Qingyang Wu, and Yong~Jae Lee.
\newblock Visual instruction tuning.
\newblock \emph{NeuIPS}, 2023.

\bibitem[Liu et~al.(2024)Liu, Chen, Cheng, Ji, Qiu, Yang, and Wang]{liu2024visual}
Minghuan Liu, Zixuan Chen, Xuxin Cheng, Yandong Ji, Ri-Zhao Qiu, Ruihan Yang, and Xiaolong Wang.
\newblock Visual whole-body control for legged loco-manipulation.
\newblock In \emph{CoRL}, 2024.

\bibitem[Liyanage et~al.(2019)Liyanage, Dia, Abduljabbar, and Bagloee]{liyanage2019flexible}
Sohani Liyanage, Hussein Dia, Rusul Abduljabbar, and Saeed~Asadi Bagloee.
\newblock Flexible mobility on-demand: An environmental scan.
\newblock \emph{Sustainability}, 2019.

\bibitem[Makoviichuk and Makoviychuk(2021)]{rl-games2021}
Denys Makoviichuk and Viktor Makoviychuk.
\newblock rl-games: A high-performance framework for reinforcement learning.
\newblock \url{https://github.com/Denys88/rl_games}, 2021.

\bibitem[Makoviychuk et~al.()Makoviychuk, Wawrzyniak, Guo, Lu, Storey, Macklin, Hoeller, Rudin, Allshire, Handa, et~al.]{makoviychuk2isaac}
Viktor Makoviychuk, Lukasz Wawrzyniak, Yunrong Guo, Michelle Lu, Kier Storey, Miles Macklin, David Hoeller, Nikita Rudin, Arthur Allshire, Ankur Handa, et~al.
\newblock Isaac gym: High performance gpu based physics simulation for robot learning.
\newblock In \emph{NeurIPS Datasets and Benchmarks}.

\bibitem[Martinez et~al.(2017)Martinez, Sitawarin, Finch, Meincke, Yablonski, and Kornhauser]{martinez2017beyond}
Mark Martinez, Chawin Sitawarin, Kevin Finch, Lennart Meincke, Alex Yablonski, and Alain Kornhauser.
\newblock Beyond grand theft auto v for training, testing and enhancing deep learning in self driving cars.
\newblock \emph{arXiv preprint arXiv:1712.01397}, 2017.

\bibitem[Masoud et~al.(2019)Masoud, Elhenawy, Almannaa, Liu, Glaser, and Rakotonirainy]{masoud2019heuristic}
Mahmoud Masoud, Mohammed Elhenawy, Mohammed~H Almannaa, Shi~Qiang Liu, Sebastien Glaser, and Andry Rakotonirainy.
\newblock Heuristic approaches to solve e-scooter assignment problem.
\newblock \emph{IEEE access}, 2019.

\bibitem[Micromobility(2020)]{micromobility2020report}
ITF~Safe Micromobility.
\newblock Report by the international transport forum oecd/itf.
\newblock In \emph{International Transport Forum: Paris, France}, 2020.

\bibitem[Midgley(2009)]{midgley2009shared}
P Midgley.
\newblock Shared smart bicycle schemes in european cities.
\newblock \emph{Global Transport Knowledge Partnership}, 2009.

\bibitem[Miki et~al.(2022)Miki, Lee, Hwangbo, Wellhausen, Koltun, and Hutter]{miki2022learning}
Takahiro Miki, Joonho Lee, Jemin Hwangbo, Lorenz Wellhausen, Vladlen Koltun, and Marco Hutter.
\newblock Learning robust perceptive locomotion for quadrupedal robots in the wild.
\newblock \emph{Science robotics}, 2022.

\bibitem[Milakis et~al.(2020)Milakis, Gedhardt, Ehebrecht, and Lenz]{milakis2020micro}
Dimitris Milakis, Laura Gedhardt, Daniel Ehebrecht, and Barbara Lenz.
\newblock Is micro-mobility sustainable? an overview of implications for accessibility, air pollution, safety, physical activity and subjective wellbeing.
\newblock \emph{Handbook of sustainable transport}, 2020.

\bibitem[Mittal et~al.(2023)Mittal, Yu, Yu, Liu, Rudin, Hoeller, Yuan, Singh, Guo, Mazhar, et~al.]{mittal2023orbit}
Mayank Mittal, Calvin Yu, Qinxi Yu, Jingzhou Liu, Nikita Rudin, David Hoeller, Jia~Lin Yuan, Ritvik Singh, Yunrong Guo, Hammad Mazhar, et~al.
\newblock Orbit: A unified simulation framework for interactive robot learning environments.
\newblock \emph{RAL}, 2023.

\bibitem[{Nvidia Corp.}(2024{\natexlab{a}})]{nvidia2024isaacsim}
{Nvidia Corp.}
\newblock Isaac sim.
\newblock \url{https://developer.nvidia.com/isaac/sim}, 2024{\natexlab{a}}.
\newblock Accessed: 2024-11.

\bibitem[{Nvidia Corp.}(2024{\natexlab{b}})]{nvidia2024omniverse}
{Nvidia Corp.}
\newblock Nvidia omniverse.
\newblock \url{https://developer.nvidia.com/omniverse}, 2024{\natexlab{b}}.
\newblock Accessed: 2024-11.

\bibitem[{Nvidia Corp.}(2024{\natexlab{c}})]{nvidia2024physx}
{Nvidia Corp.}
\newblock Physx.
\newblock \url{https://developer.nvidia.com/physx-sdk}, 2024{\natexlab{c}}.
\newblock Accessed: 2024-11.

\bibitem[Oeschger et~al.(2020)Oeschger, Carroll, and Caulfield]{oeschger2020micromobility}
Giulia Oeschger, P{\'a}raic Carroll, and Brian Caulfield.
\newblock Micromobility and public transport integration: The current state of knowledge.
\newblock \emph{Transportation Research Part D: Transport and Environment}, 2020.

\bibitem[Park et~al.(2023)Park, O'Brien, Cai, Morris, Liang, and Bernstein]{park2023generative}
Joon~Sung Park, Joseph O'Brien, Carrie~Jun Cai, Meredith~Ringel Morris, Percy Liang, and Michael~S Bernstein.
\newblock Generative agents: Interactive simulacra of human behavior.
\newblock In \emph{UIST}, 2023.

\bibitem[Puig et~al.(2018)Puig, Ra, Boben, Li, Wang, Fidler, and Torralba]{puig2018virtualhome}
Xavier Puig, Kevin Ra, Marko Boben, Jiaman Li, Tingwu Wang, Sanja Fidler, and Antonio Torralba.
\newblock Virtualhome: Simulating household activities via programs.
\newblock In \emph{CVPR}, 2018.

\bibitem[Puig et~al.(2023{\natexlab{a}})Puig, Shu, Tenenbaum, and Torralba]{puig2023nopa}
Xavier Puig, Tianmin Shu, Joshua~B Tenenbaum, and Antonio Torralba.
\newblock Nopa: Neurally-guided online probabilistic assistance for building socially intelligent home assistants.
\newblock In \emph{ICRA}, 2023{\natexlab{a}}.

\bibitem[Puig et~al.(2023{\natexlab{b}})Puig, Undersander, Szot, Cote, Yang, Partsey, Desai, Clegg, Hlavac, Min, Vondrus, Gervet, Berges, Turner, Maksymets, Kira, Kalakrishnan, Malik, Chaplot, Jain, Batra, Rai, and Mottaghi]{puig2023habitat}
Xavier Puig, Eric Undersander, Andrew Szot, Mikael~Dallaire Cote, Tsung{-}Yen Yang, Ruslan Partsey, Ruta Desai, Alexander~William Clegg, Michal Hlavac, So~Yeon Min, Vladimir Vondrus, Th{\'{e}}ophile Gervet, Vincent{-}Pierre Berges, John~M. Turner, Oleksandr Maksymets, Zsolt Kira, Mrinal Kalakrishnan, Jitendra Malik, Devendra~Singh Chaplot, Unnat Jain, Dhruv Batra, Akshara Rai, and Roozbeh Mottaghi.
\newblock Habitat 3.0: A co-habitat for humans, avatars, and robots.
\newblock In \emph{ICLR}, 2023{\natexlab{b}}.

\bibitem[Puterman(1990)]{puterman1990markov}
Martin~L Puterman.
\newblock Markov decision processes.
\newblock \emph{Handbooks in operations research and management science}, 1990.

\bibitem[Rutherford et~al.(2023)Rutherford, Ellis, Gallici, Cook, Lupu, Ingvarsson, Willi, Khan, de~Witt, Souly, et~al.]{rutherford2023jaxmarl}
Alexander Rutherford, Benjamin Ellis, Matteo Gallici, Jonathan Cook, Andrei Lupu, Gardar Ingvarsson, Timon Willi, Akbir Khan, Christian~Schroeder de Witt, Alexandra Souly, et~al.
\newblock Jaxmarl: Multi-agent rl environments in jax.
\newblock \emph{arXiv preprint arXiv:2311.10090}, 2023.

\bibitem[Savva et~al.(2019)Savva, Malik, Parikh, Batra, Kadian, Maksymets, Zhao, Wijmans, Jain, Straub, Liu, and Koltun]{savva2019habitat}
Manolis Savva, Jitendra Malik, Devi Parikh, Dhruv Batra, Abhishek Kadian, Oleksandr Maksymets, Yili Zhao, Erik Wijmans, Bhavana Jain, Julian Straub, Jia Liu, and Vladlen Koltun.
\newblock Habitat: {A} platform for embodied {AI} research.
\newblock In \emph{ICCV}, 2019.

\bibitem[Schulman et~al.(2017)Schulman, Wolski, Dhariwal, Radford, and Klimov]{schulman2017proximal}
John Schulman, Filip Wolski, Prafulla Dhariwal, Alec Radford, and Oleg Klimov.
\newblock Proximal policy optimization algorithms.
\newblock \emph{arXiv preprint arXiv:1707.06347}, 2017.

\bibitem[Shah and Levine(2022)]{shah2022viking}
Dhruv Shah and Sergey Levine.
\newblock Viking: Vision-based kilometer-scale navigation with geographic hints.
\newblock In \emph{RSS}, 2022.

\bibitem[Shaheen et~al.(2013)Shaheen, Martin, and Cohen]{shaheen2013public}
Susan Shaheen, Elliot Martin, and Adam Cohen.
\newblock Public bikesharing and modal shift behavior: a comparative study of early bikesharing systems in north america.
\newblock 2013.

\bibitem[Shaheen et~al.(2010)Shaheen, Guzman, and Zhang]{shaheen2010bikesharing}
Susan~A Shaheen, Stacey Guzman, and Hua Zhang.
\newblock Bikesharing in europe, the americas, and asia: past, present, and future.
\newblock \emph{Transportation research record}, 2010.

\bibitem[Shen et~al.(2021)Shen, Xia, Li, Mart{\'\i}n-Mart{\'\i}n, Fan, Wang, P{\'e}rez-D’Arpino, Buch, Srivastava, Tchapmi, Micael, Kent, Josiah, Li, and Silvio]{shen2021igibson}
Bokui Shen, Fei Xia, Chengshu Li, Roberto Mart{\'\i}n-Mart{\'\i}n, Linxi Fan, Guanzhi Wang, Claudia P{\'e}rez-D’Arpino, Shyamal Buch, Sanjana Srivastava, Lyne Tchapmi, Tchapmi Micael, Vainio Kent, Wong Josiah, Fei-Fei Li, and Savarese Silvio.
\newblock igibson 1.0: a simulation environment for interactive tasks in large realistic scenes.
\newblock In \emph{IROS}, 2021.

\bibitem[Short and Adams(2017)]{short2017procedural}
Tanya Short and Tarn Adams.
\newblock \emph{Procedural generation in game design}.
\newblock CRC Press, 2017.

\bibitem[Smith et~al.(2023)Smith, Kew, Li, Luu, Peng, Ha, Tan, and Levine]{smith2023learning}
Laura~M. Smith, J.~Chase Kew, Tianyu Li, Linda Luu, Xue~Bin Peng, Sehoon Ha, Jie Tan, and Sergey Levine.
\newblock Learning and adapting agile locomotion skills by transferring experience.
\newblock In \emph{RSS}, 2023.

\bibitem[Sorokin et~al.(2022)Sorokin, Tan, Liu, and Ha]{sorokin2022learning}
Maks Sorokin, Jie Tan, C~Karen Liu, and Sehoon Ha.
\newblock Learning to navigate sidewalks in outdoor environments.
\newblock \emph{RAL}, 2022.

\bibitem[Szot et~al.(2021)Szot, Clegg, Undersander, Wijmans, Zhao, Turner, Maestre, Mukadam, Chaplot, Maksymets, Gokaslan, Vondrus, Dharur, Meier, Galuba, Chang, Kira, Koltun, Malik, Savva, and Batra]{szot2021habitat}
Andrew Szot, Alexander Clegg, Eric Undersander, Erik Wijmans, Yili Zhao, John~M. Turner, Noah Maestre, Mustafa Mukadam, Devendra~Singh Chaplot, Oleksandr Maksymets, Aaron Gokaslan, Vladimir Vondrus, Sameer Dharur, Franziska Meier, Wojciech Galuba, Angel~X. Chang, Zsolt Kira, Vladlen Koltun, Jitendra Malik, Manolis Savva, and Dhruv Batra.
\newblock Habitat 2.0: Training home assistants to rearrange their habitat.
\newblock In \emph{NeuIPS}, 2021.

\bibitem[Tiwari(2019)]{tiwari2019micro}
A Tiwari.
\newblock Micro-mobility: the next wave of urban transportation in india.
\newblock \emph{YS Journal, January}, 2019.

\bibitem[Tran et~al.(2015)Tran, Ovtracht, and d’Arcier]{tran2015modeling}
Tien~Dung Tran, Nicolas Ovtracht, and Bruno~Faivre d’Arcier.
\newblock Modeling bike sharing system using built environment factors.
\newblock \emph{Procedia Cirp}, 2015.

\bibitem[Treiber et~al.(2000)Treiber, Hennecke, and Helbing]{treiber2000congested}
Martin Treiber, Ansgar Hennecke, and Dirk Helbing.
\newblock Congested traffic states in empirical observations and microscopic simulations.
\newblock \emph{Physical review E}, 62\penalty0 (2):\penalty0 1805, 2000.

\bibitem[Tsoi et~al.(2022)Tsoi, Xiang, Yu, Sohn, Schwartz, Ramesh, Hussein, Gupta, Kapadia, and V{\'a}zquez]{tsoi2022sean}
Nathan Tsoi, Alec Xiang, Peter Yu, Samuel~S Sohn, Greg Schwartz, Subashri Ramesh, Mohamed Hussein, Anjali~W Gupta, Mubbasir Kapadia, and Marynel V{\'a}zquez.
\newblock Sean 2.0: Formalizing and generating social situations for robot navigation.
\newblock \emph{RAL}, 2022.

\bibitem[{U.S. Department of Transportation}()]{nhts2023}
{U.S. Department of Transportation}.
\newblock Fatality analysis reporting system (fars).
\newblock \url{https://www.nhtsa.gov/research-data/fatality-analysis-reporting-system-fars}.

\bibitem[{U.S. Department of Transportation}(2024)]{dot}
{U.S. Department of Transportation}.
\newblock Transportation reports and publications.
\newblock \url{https://www.transportation.gov/}, 2024.
\newblock Accessed: 2024-11.

\bibitem[{U.S. Federal Highway Administration}(2017)]{nhts}
{U.S. Federal Highway Administration}.
\newblock National household travel survey (nhts).
\newblock \url{https://nhts.ornl.gov/}, 2017.
\newblock Accessed: 2024-11.

\bibitem[Van Den~Berg et~al.(2011)Van Den~Berg, Guy, Lin, and Manocha]{van2011reciprocal}
Jur Van Den~Berg, Stephen~J Guy, Ming Lin, and Dinesh Manocha.
\newblock Reciprocal n-body collision avoidance.
\newblock In \emph{ISRR}, 2011.

\bibitem[Wang et~al.(2024)Wang, Chen, Huang, Ben, Wang, Mi, Huang, Zhao, Chen, Yang, et~al.]{wang2024grutopia}
Hanqing Wang, Jiahe Chen, Wensi Huang, Qingwei Ben, Tai Wang, Boyu Mi, Tao Huang, Siheng Zhao, Yilun Chen, Sizhe Yang, et~al.
\newblock Grutopia: Dream general robots in a city at scale.
\newblock \emph{arXiv preprint arXiv:2407.10943}, 2024.

\bibitem[White and Clarke(2020)]{white2020urban}
Gary White and Siobhan Clarke.
\newblock Urban intelligence with deep edges.
\newblock \emph{IEEE Access}, 2020.

\bibitem[Wu et~al.(2023)Wu, Zhang, Fu, Wang, Ren, Pan, Wu, Yang, Wang, Qian, Lin, and Liu]{wu2023omniobject3d}
Tong Wu, Jiarui Zhang, Xiao Fu, Yuxin Wang, Jiawei Ren, Liang Pan, Wayne Wu, Lei Yang, Jiaqi Wang, Chen Qian, Dahua Lin, and Ziwei Liu.
\newblock Omniobject3d: Large-vocabulary 3d object dataset for realistic perception, reconstruction and generation.
\newblock In \emph{CVPR}, 2023.

\bibitem[Wu et~al.(2024)Wu, He, He, Wang, Duan, Liu, Li, and Zhou]{wu2024metaurban}
Wayne Wu, Honglin He, Jack He, Yiran Wang, Chenda Duan, Zhizheng Liu, Quanyi Li, and Bolei Zhou.
\newblock Metaurban: An embodied ai simulation platform for urban micromobility.
\newblock \emph{arXiv preprint arXiv:2407.08725}, 2024.

\bibitem[Xie et~al.(2024)Xie, Chen, Hong, and Liu]{xie2024citydreamer}
Haozhe Xie, Zhaoxi Chen, Fangzhou Hong, and Ziwei Liu.
\newblock Citydreamer: Compositional generative model of unbounded 3d cities.
\newblock In \emph{CVPR}, 2024.

\bibitem[Yang et~al.(2020)Yang, Ma, Wang, Cai, Xie, and Yang]{yang2020safety}
Hong Yang, Qingyu Ma, Zhenyu Wang, Qing Cai, Kun Xie, and Di Yang.
\newblock Safety of micro-mobility: Analysis of e-scooter crashes by mining news reports.
\newblock \emph{Accident Analysis \& Prevention}, 2020.

\bibitem[Yang et~al.(2023)Yang, Cai, Mei, Liu, Chen, Xiao, Wei, Qing, Wei, Dai, Wu, Qian, Lin, Liu, and Yang]{yang2023synbody}
Zhitao Yang, Zhongang Cai, Haiyi Mei, Shuai Liu, Zhaoxi Chen, Weiye Xiao, Yukun Wei, Zhongfei Qing, Chen Wei, Bo Dai, Wayne Wu, Chen Qian, Dahua Lin, Ziwei Liu, and Lei Yang.
\newblock Synbody: Synthetic dataset with layered human models for 3d human perception and modeling.
\newblock In \emph{ICCV}, 2023.

\bibitem[Yokoyama et~al.(2024)Yokoyama, Ramrakhya, Das, Batra, and Ha]{yokoyama2024hm3d}
Naoki Yokoyama, Ram Ramrakhya, Abhishek Das, Dhruv Batra, and Sehoon Ha.
\newblock Hm3d-ovon: A dataset and benchmark for open-vocabulary object goal navigation.
\newblock \emph{arXiv preprint arXiv:2409.14296}, 2024.

\bibitem[Zhang et~al.(2024)Zhang, Zhou, Wang, Luo, Wang, Li, Yin, Zhang, and Peng]{zhang2024cityx}
Shougao Zhang, Mengqi Zhou, Yuxi Wang, Chuanchen Luo, Rongyu Wang, Yiwei Li, Xucheng Yin, Zhaoxiang Zhang, and Junran Peng.
\newblock Cityx: Controllable procedural content generation for unbounded 3d cities.
\newblock \emph{arXiv preprint arXiv:2407.17572}, 2024.

\end{thebibliography}
}

\clearpage

\appendix
\begin{abstract}

In the appendix, we present more details of this work.
In Section~\ref{sec:urbansim_visualization}, we illustrate samples of scenes for Urban Locomotion, Navigation, and Traverse tasks.
In Section~\ref{sec:urbansim_simulator}, we introduce the design details of \texttt{URBAN-SIM} platform.
In Section~\ref{sec:performance_benchmark}, we construct extensive performance benchmarks of \texttt{URBAN-SIM}.
In Section~\ref{sec:approach}, we give details and discuss the properties of the human-AI shared autonomous approach.
In Section~\ref{sec:exp_detail}, we elaborate on implementation details in the experiments of \texttt{URBAN-BENCH} and evaluate the necessity of foundational tasks.
In Section~\ref{sec:data_sheet}, we provide the datasheet of \texttt{URBAN-SIM}.
In Section~\ref{sec:discussion}, we discuss the impacts, sim-to-real generalization, limitations, and future directions of this work.

\end{abstract}

{
\startcontents[sections]
\printcontents[sections]{l}{1}{\setcounter{tocdepth}{3}}
}
\section{URBAN-SIM Visualization}
\label{sec:urbansim_visualization}

In this section, we will give visualization results of the training and evaluation scenes for the three skills in autonomous micromobility -- Urban Locomotion, Urban Navigation, and Urban Traverse.

\subsection{Urban Locomotion Scene Samples}

For the training and testing in Urban Locomotion, we construct four types of terrains, flat, slope, stair, and rough surfaces, for four foundational tasks. For each type of terrain, we sample different difficulty parameters to construct scenes. Note that, based on the terrain generation module provided in \texttt{URBAN-SIM}, users can easily define new urban locomotion tasks to meet specific demands, such as ``climb high steps'', and "jump over the gaps between slabs".

\begin{figure*}[t!]
    \centering
    \includegraphics[width=1\linewidth]{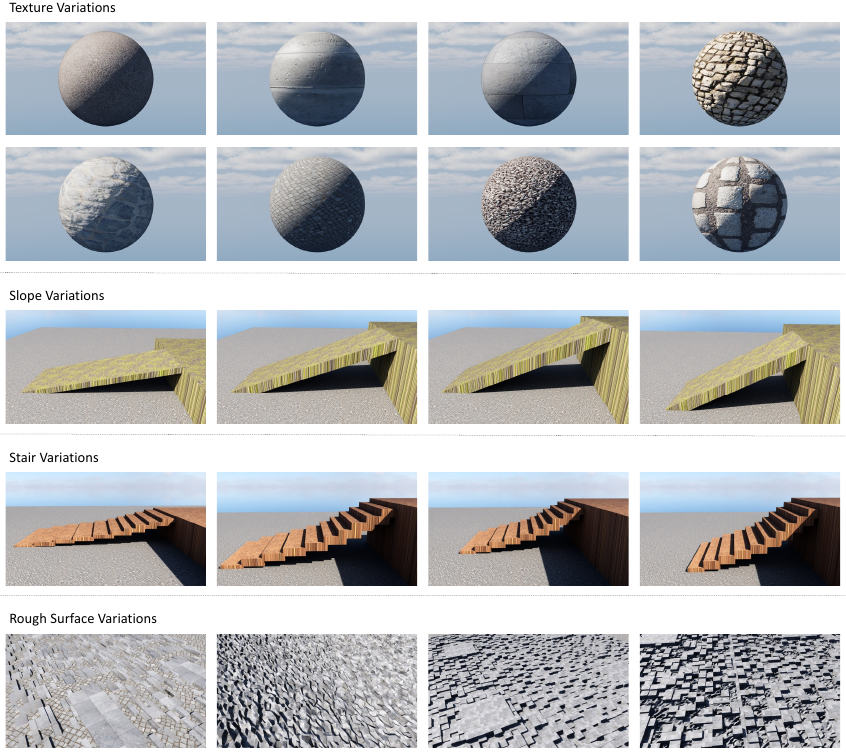}
    \caption{\textbf{Samples of different settings of terrains.}
    }
    \label{fig:loco}
\end{figure*}

We demonstrate the terrains generated by the terrain generation module below.
Figure~\ref{fig:loco} (1st row) shows sampled 8 types of textures that can be set with varied friction coefficients.
Figure~\ref{fig:loco} (2nd row) shows sampled 4 dip angles of inclined surfaces.
Figure~\ref{fig:loco} (3rd row) shows sampled 4 heights and widths of the stairs.
Figure~\ref{fig:loco} (4th row) shows sampled 4 bumpiness of the ground of rough surfaces.

\subsection{Urban Navigation Scene Samples}

For the training and testing in Urban Navigation, we construct three types of situations, unobstructed ground, static obstacles, and dynamic obstacles, for three foundational tasks. For each type of situation, we sample different difficulty parameters to construct scenes. The urban navigation scenes have been equipped with mixed terrains randomly sampled from four different types of terrains used in urban locomotion tasks. The scenes for urban navigation tasks are built progressively: scenes with static obstacles are created based on unobstructed ground, and scenes with dynamic obstacles are built upon static obstacles.
Note that, based on the hierarchical urban generation pipeline provided in \texttt{URBAN-SIM}, users can easily define new urban navigation tasks to meet specific demands, such as ``find path to curb ramp'', and ``go cross the intersection''.

\paragraph{NavClear scenes.}

\begin{figure*}[h!]
    \centering
    \includegraphics[width=1\linewidth]{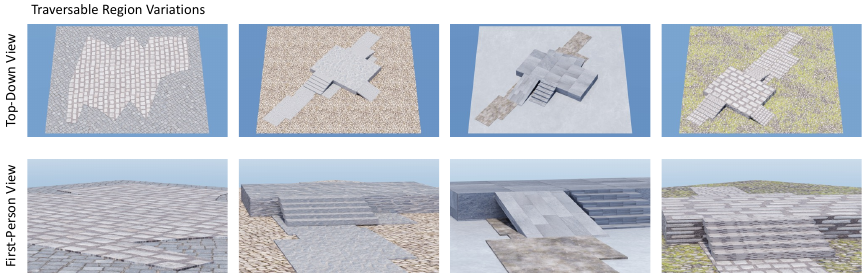}
    \caption{\textbf{Samples of different settings of traversable regions.}
    }
    \label{fig:nav_clear}
\end{figure*}

For the scenes used to train and test the NavClear task, we categorize difficulty levels based on the types and shapes of traversable regions. We sample 4 different settings of traversable regions in Figure~\ref{fig:nav_clear}, and we define the center part of one scene as a traversable region. The first Row is the top-down view, while the second row is the first-person view.

\paragraph{NavStatic scenes.}

\begin{figure*}[h!]
    \centering
    \includegraphics[width=1\linewidth]{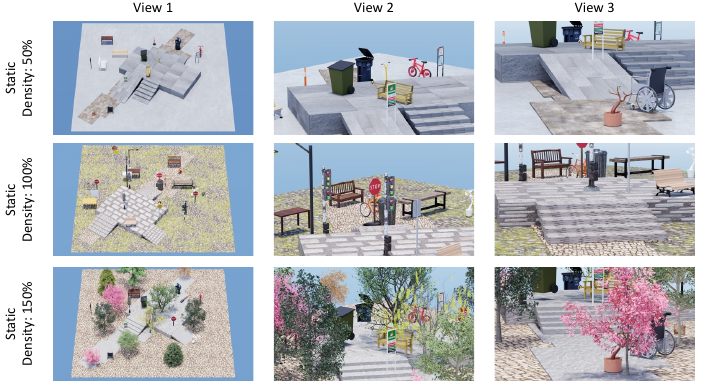}
    \caption{\textbf{Samples of different densities of static obstacles.}
    }
    \label{fig:nav_static}
\end{figure*}

For the scenes used to train and test the NavStatic task, we categorize difficulty levels based on the density of static obstacles, such as benches, trash bins, and advertising boards. We sample 3 different densities (50$\%$, 100$\%$, and 150$\%$) of static obstacles and show 3 views for each density in Figure~\ref{fig:nav_static}.

\paragraph{NavDynamic scenes.}

\begin{figure*}[h!]
    \centering
    \includegraphics[width=1\linewidth]{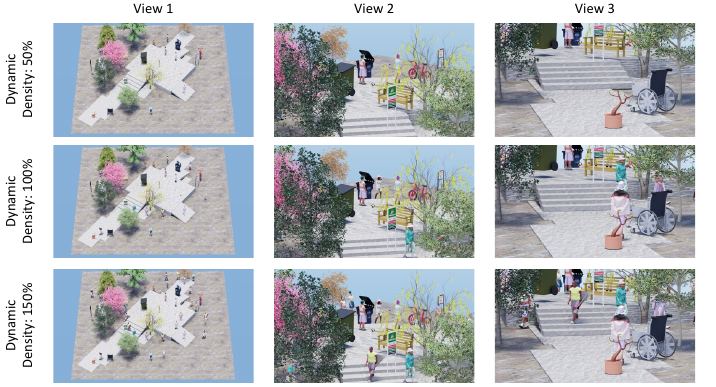}
    \caption{\textbf{Samples of different densities of dynamic obstacles.}
    }
    \label{fig:nav_dynamic}
\end{figure*}

For the scenes used to train and test the NavDynamic task, we categorize difficulty levels based on the density of dynamic obstacles, such as pedestrians, cyclists, and scooter riders. We sample 3 different densities (50$\%$, 100$\%$, and 150$\%$) of dynamic obstacles and show 3 views for each density in Figure~\ref{fig:nav_dynamic}. In these results, we randomly sample the static object with a density of 100$\%$.

\paragraph{Scene scales.}

\begin{figure*}[h!]
    \centering
    \includegraphics[width=1\linewidth]{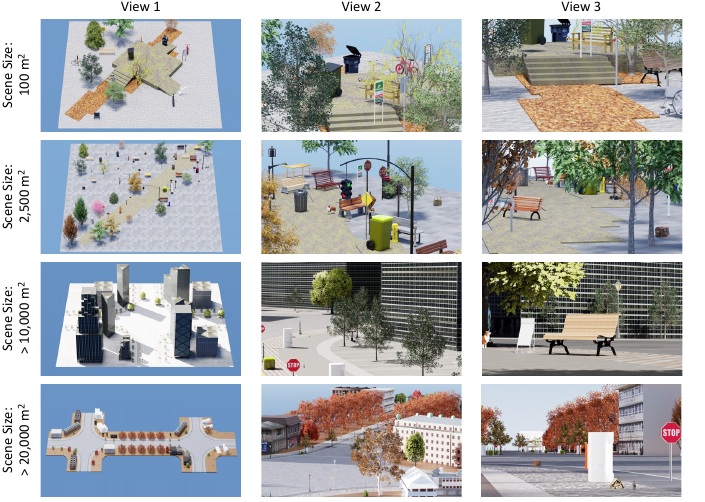}
    \caption{\textbf{Samples of different scene scales.}
    }
    \label{fig:scene_scale}
\end{figure*}

For each task in urban navigation, users can further choose the scene scale as they want, from street corner size (25 $m^2$) to city scale (200,000 $m^2$). For areas smaller than 10,000 $m^2$, users can create regular rectangular regions. In Figure~\ref{fig:scene_scale}, we sample square regions with three different side lengths (10 $m$, 50 $m$, and 100 $m$) and present five variations for each size.

For areas larger than 10,000 $m^2$, users have the option to use regular rectangles or a combination of street blocks. In Figure~\ref{fig:scene_scale} (the 4th column), we showcase five large-scale scenes featuring various combinations of street blocks. Figure~\ref{fig:scene_scale} (Bottom), we further sample ego-view images from these scenes.

\subsection{Urban Traverse Scene Samples}

For the testing in Urban Traverse, we construct two large-scale test scenes: Urban-Tra-Standard and Urban-Tra-City. These two scenes are generated by the hierarchical urban generation system of \texttt{URBAN-SIM} and then adjusted manually to better fit the specific evaluation needs. For example, we manually established two traversable routes for a wheeled robot (with no stairs and no extremely uneven terrain) and a legged robot of the same length for a fair comparison. Note that, empowered by the UI of Omniverse~\cite{nvidia2024omniverse}, users can easily modify the scenes generated by our pipeline or directly manually build scenes with urban assets provided in \texttt{URBAN-SIM}. This flexibility enables the evaluation of different aspects of robot capabilities with specifically designed scenes.

\paragraph{Urban-Tra-Standard.}

\begin{figure*}[h!]
    \centering
    \includegraphics[width=1\linewidth]{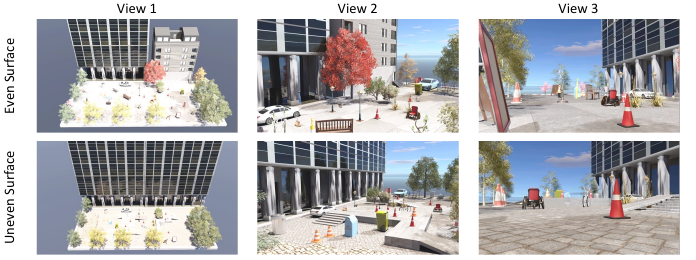}
    \caption{\textbf{Samples of different scenarios in the standard testing environment (Urban-Tra-Standard).}
    }
    \label{fig:sample_standard}
\end{figure*}

This scene is used for standardized testing of the urban traverse task, covering six scenarios and carefully considering the traversability of different robots.
The six scenarios include two parts:
1) Even surfaces with unobstructed ground, static obstacles, and dynamic obstacles.
2) Uneven surfaces with unobstructed ground, static obstacles, and dynamic obstacles.
The overall scene measures 120 $m$ in length and 10 $m$ in width, while each individual scenario is 20 $m$ long and 10 $m$ wide.
We show a sampled scenario for even and uneven surfaces, respectively, in Figure~\ref{fig:sample_standard}. Each row shows 3 views: two side views and a first-person view.

\paragraph{Urban-Tra-City}

\begin{figure*}[h!]
    \centering
    \includegraphics[width=1\linewidth]{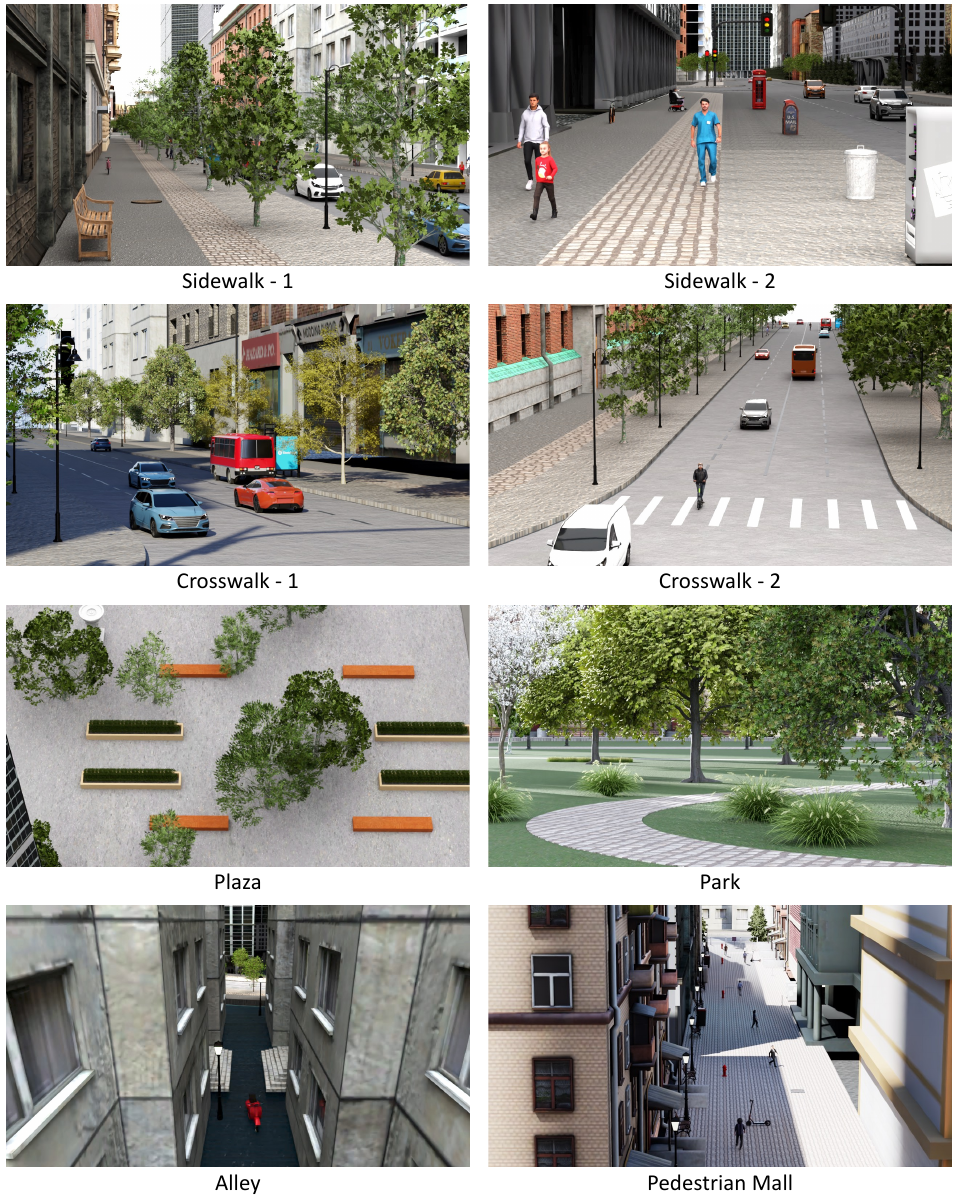}
    \caption{\textbf{Samples of different functional zones in the city-scale testing environment (Urban-Tra-City).}
    }
    \label{fig:sample_city}
\end{figure*}

This scene is used for the qualitative testing of the urban traverse task in real-world scenarios, covering different functional zones in a city, such as sidewalks, crosswalks, and parks. This scene features vivid and realistic urban scenarios, including diverse layouts, complex terrains, city facilities, traffic flows, and movements of dense pedestrians. These elements present an intricate evaluation environment for robots in the urban traverse task. The overall scene measures 1,200 $m$ in width and 1,200 $m$ in length, resulting in an area of 1,440,000 $m^2$.
We demonstrate 6 different urban function zones sampled from Urban-Tra-City in Figure~\ref{fig:sample_city}, \ie, sidewalk, crosswalk, plaza, park, alley, and pedestrian mall.

\clearpage
\section{URBAN-SIM Design Details}
\label{sec:urbansim_simulator}

In this section, we will introduce the design details of \texttt{URBAN-SIM}, including the urban assets and supported robots.

\subsection{Static Assets}

\begin{figure*}[h!]
    \centering
    \includegraphics[width=1\linewidth]{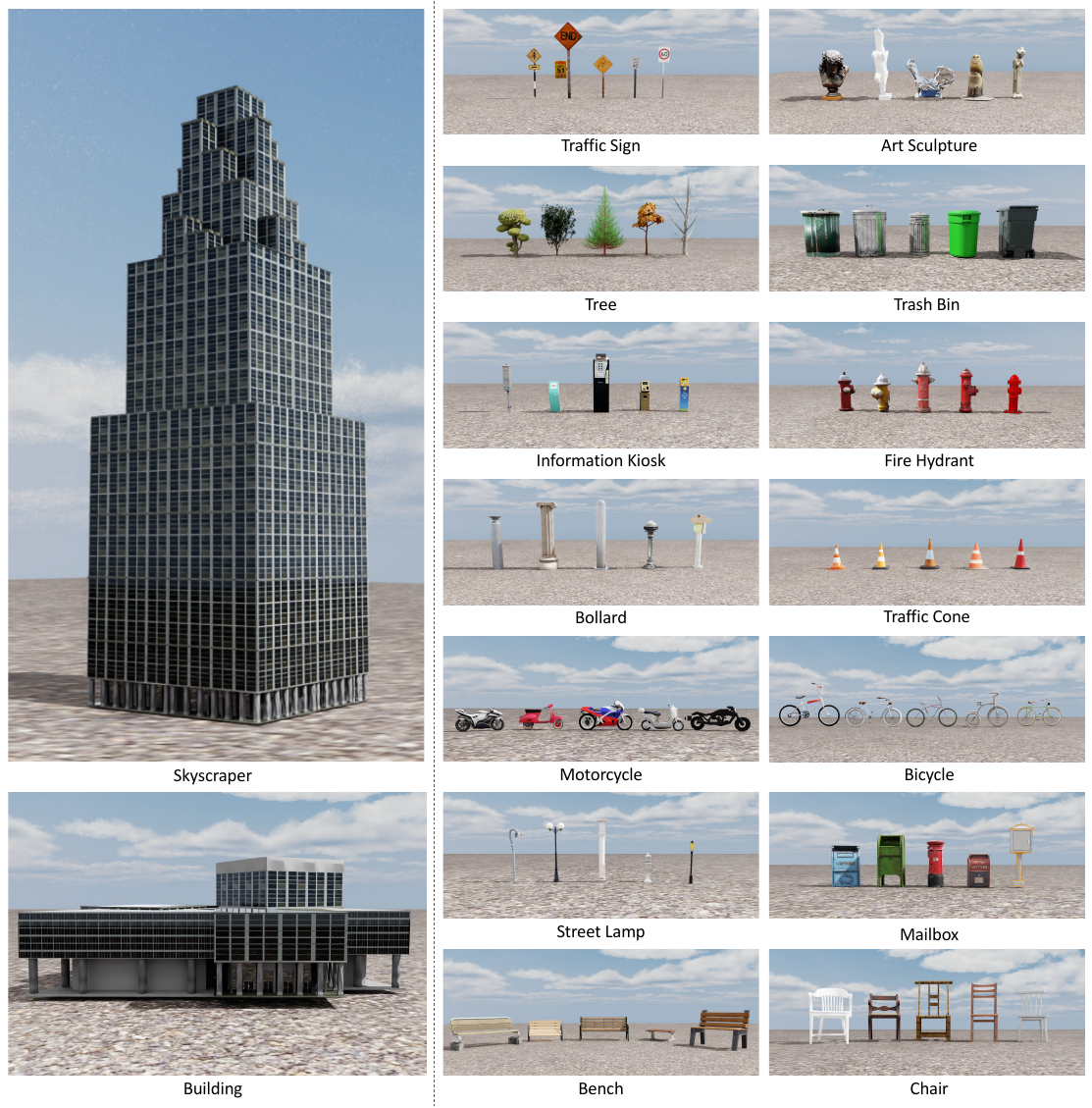}
    \caption{\textbf{Samples of static assets in urban spaces.}
    }
    \label{fig:static_asset}
\end{figure*}

We collect static assets covering urban facilities like buildings, trees, mailboxes, bus stops, and telephone booths, from varied sources, such as large-scale object dataset~\cite{wu2023omniobject3d,deitke2024objaverse}, simulation environments~\cite{dosovitskiy2017carla}, and purchase from high-quality 3D asset stores~\cite{UnityAssetStore,UnrealMarketplace}. We finally constructed a static repository of 15,000 assets, 
In Figure~\ref{fig:static_asset}, we show 15 representative categories in urban spaces.

\subsection{Dynamic Assets}

\begin{figure*}[h!]
    \centering
    \includegraphics[width=1\linewidth]{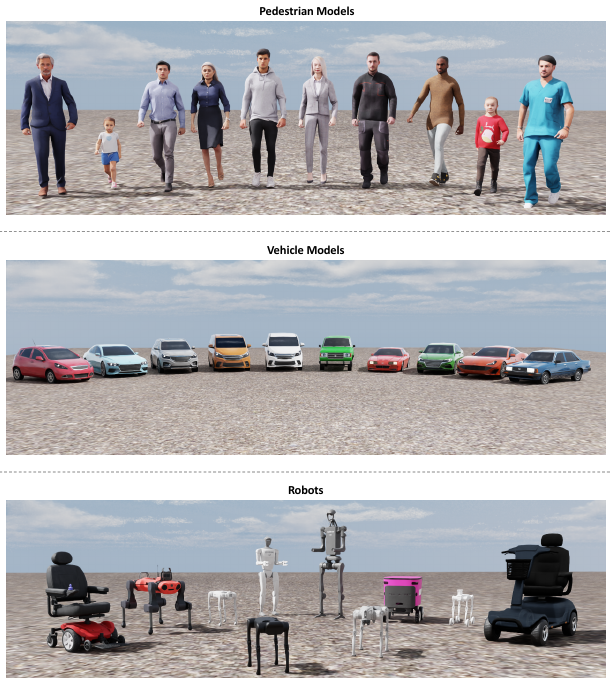}
    \caption{\textbf{Samples of dynamic assets and robots in urban spaces.}
    }
    \label{fig:dynamic_asset_and_robot}
\end{figure*}

We collect dynamic assets covering urban participants like pedestrians, vehicles, scooters, and bicycles from varied sources, such as 3D human dataset~\cite{yang2023synbody} and purchase from high-quality 3D asset stores~\cite{UnrealMarketplace,Renderpeople}. All of the human assets are rigged and can be animated by motion sequences with different movements. We finally constructed a dynamic repository of 1,620 assets.
We show samples of pedestrian and vehicle models in Figure~\ref{fig:dynamic_asset_and_robot} (Top and Middle).

\subsection{Robots}

We support the training of robots with standard URDF files. We have conducted function tests of 10 robots across 4 categories: wheeled robot: COCO delivery robot, an electric wheelchair, a mobility scooter; quadruped robot: Anymal-C, Unitree A1, Unitree Go1, Unitree Go2; wheeled-legged robot: Unitree B2-W; and humanoid robot: Unitree G1, Unitree H1. Note that, it is simple to import other new robots using the interface provided by \texttt{URBAN-SIM}.
We show samples of the supported robots in Figure~\ref{fig:dynamic_asset_and_robot} (Bottom).

\clearpage
\section{URBAN-SIM Performance Benchmark}
\label{sec:performance_benchmark}

\subsection{Settings}

We benchmark the speed of URBAN-SIM on navigation tasks under varying settings of scenes, including sizes of scenes, different numbers of objects in the scene, and different robots.
All tests are conducted on a single Nvidia L40S GPU with 46 GB of memory. We report results on a single environment and 16, 64, 128, and 256 parallel environments with a Reinforcement Learning framework RL Games \cite{rl-games2021}.
We sample random actions for 1,000 steps per agent and report average and standard error results across 10 runs. Our small scene is 100 $m^2$ in size with 4 objects, the medium scene is 400 $m^2$ in size with 8 objects, and the large scene is 2,500 $m^2$ in size with 16 objects.
In all benchmarks, we render RGBD images with the resolution 128 $\times$ 128 for agents.

\subsection{Single Environment Performance}

\begin{figure*}[h!]
    \centering
    \includegraphics[width=1\linewidth]{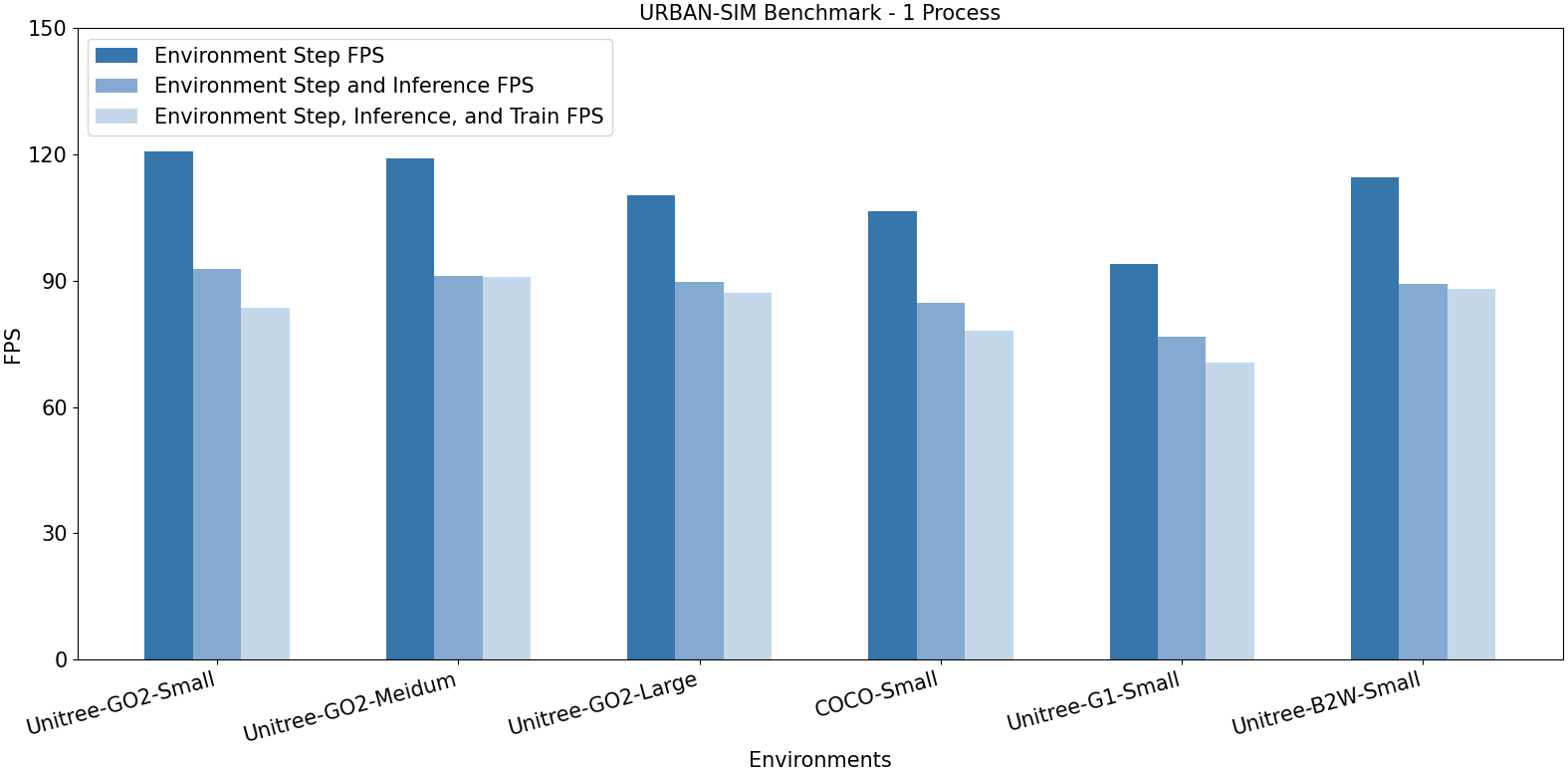}
    \caption{\textbf{Performance of 1 process for different settings of scene and robot.}
    }
    \label{fig:performance_1_process}
\end{figure*}

As shown in Figure~\ref{fig:performance_1_process}, the environment step performance ranges between 94 and 120 FPS, depending on the specific settings.
Interestingly, variations in scene size and the number of objects have minimal impact on performance speed, remaining consistent within the range of 110 to 120 $fps$.
Performance differences are more pronounced when comparing robot types.
The wheeled-legged robot (Unitree B2W) and humanoid robot (Unitree G1) demonstrate lower FPS compared to the quadruped robot (Unitree Go2) (114 vs. 94 vs. 120), primarily due to the higher number of joints and greater complexity in their skeletal models.
Surprisingly, the wheeled robot (COCO) also performs slightly worse than the quadruped robot, potentially attributed to differences in computational demands for simulating wheel-ground interactions and dynamics.
Across all settings, we observe a gradual decline in FPS when incorporating inference or training processes. This reduction is expected and is caused by the additional computational workload introduced by model execution and increased interactions within the environment.

\subsection{Multiple Environments Performance}

\begin{figure*}[h!]
    \centering
    \includegraphics[width=1\linewidth]{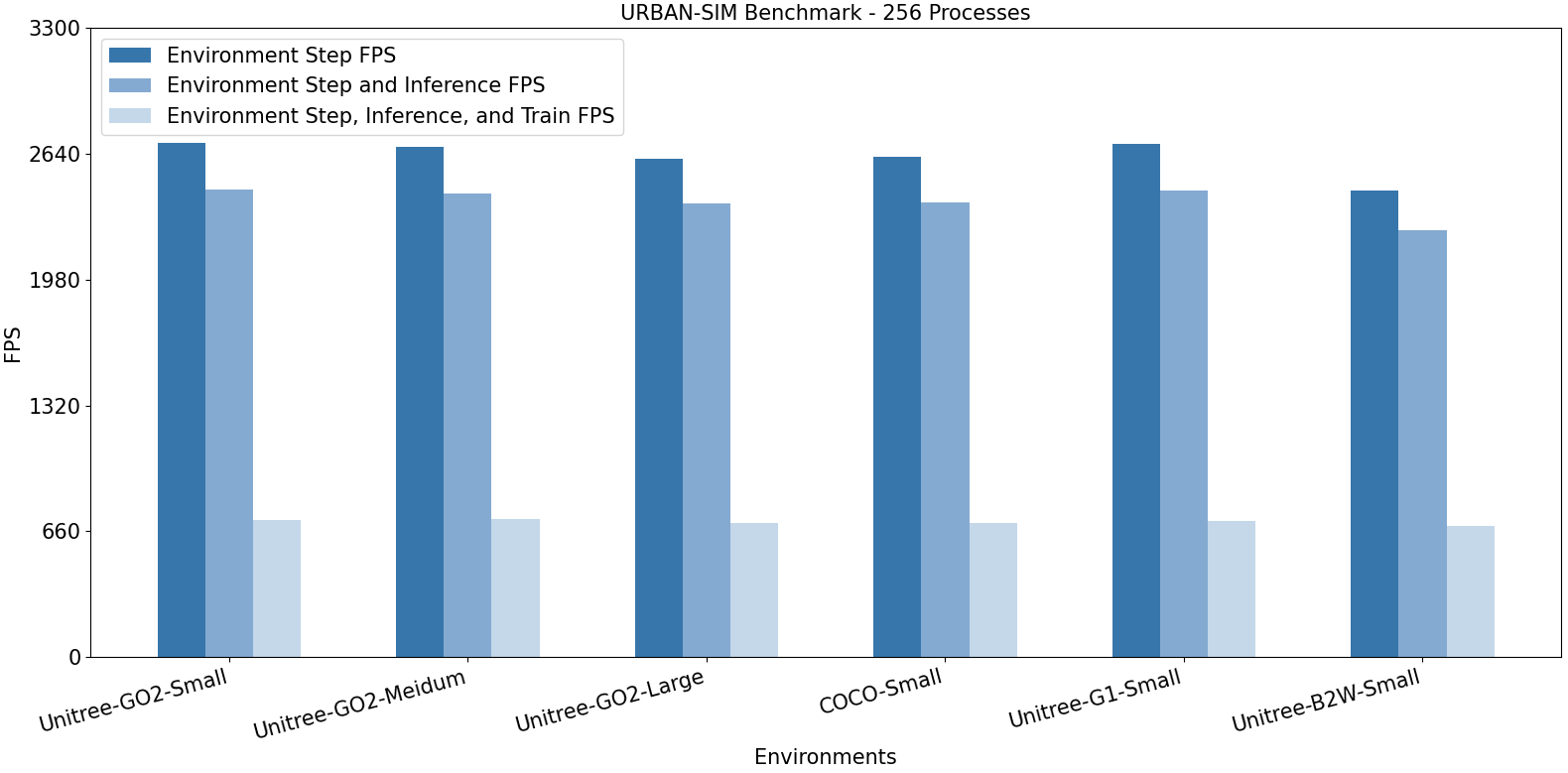}
    \caption{\textbf{Performance of 256 processes for different settings of scene and robot.}
    }
    \label{fig:performance_256_process}
\end{figure*}

As shown in Figure~\ref{fig:performance_256_process}, parallelizing up to 256 environments on a single GPU achieves impressive scalability, with performance ranging from 2,300 to 2,600 FPS.
The gap between "Environment Step FPS" and "Environment Step, Inference, and Train FPS" becomes larger compared to the single environment setting. This is likely due to the added computational overhead introduced by simultaneously handling inference and training across multiple environments.

\begin{figure}[h!]
    \centering
    \includegraphics[width=1\linewidth]{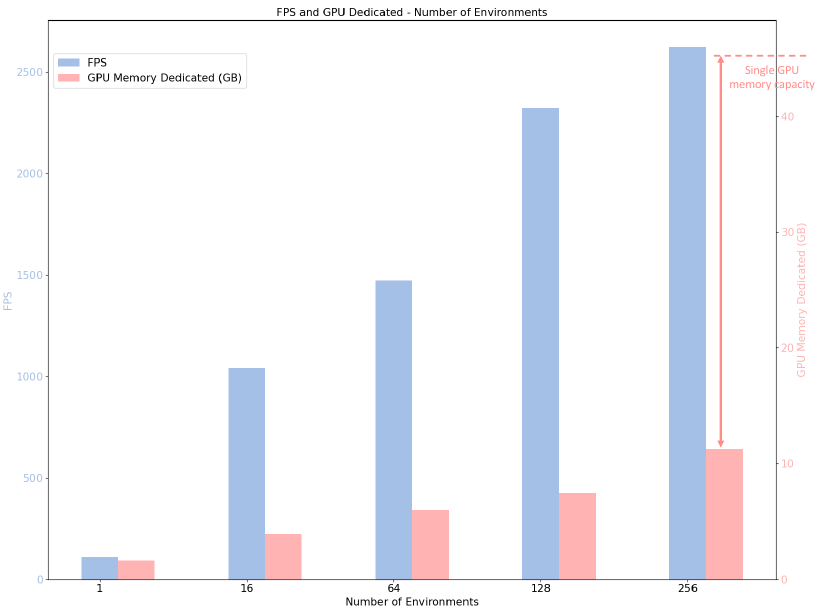}
    \caption{\textbf{FPS and GPU usage changing with the increasing number of environments.}
    }
    \label{fig:performance_w_gpu}
\end{figure}

In Figure~\ref{fig:performance_w_gpu}, we further report the FPS and GPU memory usage as the number of environments increases from 1 to 256. FPS scales significantly from 100 to 2620 $fps$ with the number of parallel environments increasing. Remarkably, GPU memory usage grows only slightly, from 1.6 to 11.2, with 256 environments occupying just 11.2 GB of the available 46 GB of memory.

These results highlight the scalability and efficiency of the \texttt{URBAN-SIM} platform in supporting large-scale robot training in diverse and complex urban environments.

\section{Human-AI Shared Autonomous Approach}
\label{sec:approach}

\begin{figure*}[h!]
    \centering
    \includegraphics[width=1\linewidth]{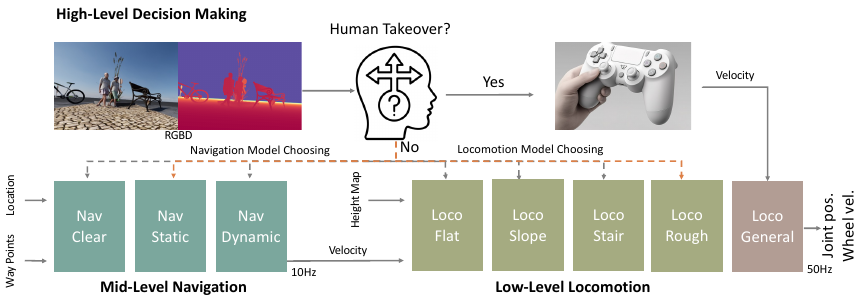}
    \caption{\textbf{Human-AI shared autonomous approach.}
    }
    \label{fig:shared_autonomy_framework}
\end{figure*}

We propose a human-AI shared autonomous approach as a pilot to address this task, combining AI capabilities with human interventions.
As illustrated in Figure~\ref{fig:shared_autonomy_framework}, in this approach, we structure the robot control into three layers: high-level decision-making, mid-level navigation, and low-level locomotion. With the layered architecture, we decompose the complex urban traverse task into a series of subtasks, with AI managing routine tasks in mid-level and low-level, and humans making high-level decisions and intervening in risky situations.
The core idea is to decompose the complex urban traverse task into a series of subtasks, with AI managing routine tasks and humans making high-level decisions or intervening in risky situations. 

We set a series of decision points along the route. At each decision point, humans assess the environment to decide whether to \textit{take control} for the next interval if the situation requires careful handling (\eg, near children) or allow the robot to execute selected navigation and locomotion models autonomously if the environment is low-risk.
This approach supports a flexible transition between human and AI control. Humans can manage the entire process if needed, while AI can manage the entire operation using an extra rule-based/AI-based decision model to direct the dispatch of urban navigation and locomotion models.
Under the shared autonomy paradigm, this approach possesses two key properties -- \textbf{Stretchability} from human to AI and \textbf{Generality} across robots. We will discuss these two properties in detail below.

\paragraph{Stretchability.} Stretchability is the ability to transition between highly human control and highly AI autonomy freely, allowing for the achievement of optimal performance in efficiency and safety based on task complexity and environmental conditions.

\begin{figure}[h!]
    \centering
    \includegraphics[width=1\linewidth]{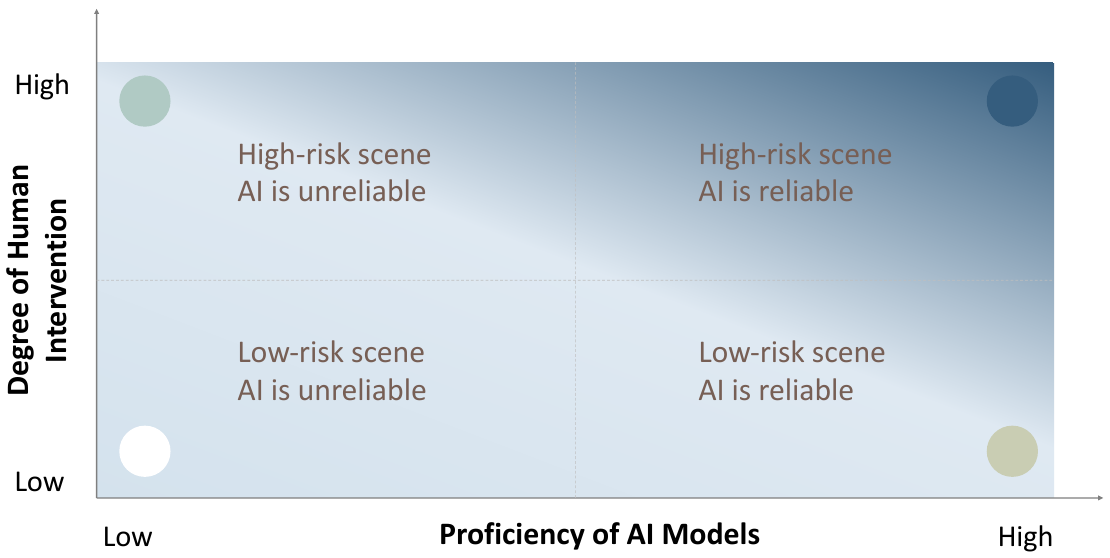}
    \caption{\textbf{Stretchability of the human-AI shared autonomous approach.}
    }
    \label{fig:stretchability}
\end{figure}

We illustrated the basic implementation logic of stretchability underlying our approach in Figure~\ref{fig:stretchability}. The stretchability comes from two functions in our framework:
1) Steerability of the degree of human intervention -- the decision-making frequency and intervention methods. We can choose the interval of decision-making points based on different conditions, such as moving distance and time duration. The denser the decision-making points are, the higher the degree of human intervention is. In addition, we can choose the intervention methods if the decision-making results need humans involved, such as directly taking over control, providing waypoints, or overseeing the behavior of agents.
2) Steerability of proficiency of AI models -- model versatility and granularity of primitive policies. We can define the skills of AI models, such as decision-making, navigation, and locomotion. For each skill, we can further freely determine the granularity of primitive policies. For example, locomotion models can be finely segmented to handle specific terrains such as steps and cobblestones.

Based on the stretchability of our approach, we can transition freely into the two-dimensional space spanning with human and AI control, as shown in Figure~\ref{fig:stretchability}.
We can adjust the degree of human intervention based on the risk degrees (the showing of children, elderly people, and heavy traffic). For example, a low-risk scene (\lowlow \highlow) could be a clear, flat sidewalk with no humans, whereas a high-risk scene (\lowhigh \highhigh) could be facing heavy traffic and a group of children playing near the intersection.
We can adjust the proficiency of AI models based on the reliance on AI's assistance (the ability of AI to solve the problem). For example, AI is reliable (\highlow \highhigh) when the models are versatile and have a high success rate across different conditions, such as the primitive locomotion and navigation models that are robust in their specific domains (as the ones in our benchmark).
Whereas AI is unreliable (\lowlow \lowhigh) when the models have a low success rate even under simple conditions, such as locomotion and navigation models that are trained with simple mixed situations.

\paragraph{Generality.} Generality is an ability generally feasible for different robots with various mechanical structures, allowing easy adaptation to robots with minimal changes.
The core advantage lies in the consistent high-level decision-making procedure, which remains unchanged across platforms, whether for wheeled robots, legged robots, or other robotic systems.
To support a new robot, the framework reuses the same decision architecture while only requiring retraining or fine-tuning of the navigation and locomotion models. For instance, when introducing a new robot, we retrain specific models to handle its unique movement and environment interaction — such as wheel-legged robots traversing flat surfaces or legged robots traversing uneven terrain. This modularity in model design makes the framework highly adaptable to any robot with different application scenarios -- individual travel or parcel delivery.

\clearpage
\section{URBAN-BENCH Experimental Details}
\label{sec:exp_detail}

\subsection{Data}
\label{sec:bench_data}

We construct 4 datasets: Urban-Nav, Urban-Loc, Urban-Tra-Standard, and Urban-Tra-City. All of the data will be released. We elaborate on the details below.

\noindent
\underline{\textit{1) Urban-Nav.}} To benchmark the urban navigation tasks, we generate 3 subsets with different environmental conditions.
1) Urban-Nav-1: scenes with the ground separated into traversable (such as trails or sidewalks) and untraversable (such as shrubs or roadways) areas.
2) Urban-Nav-2: scenes with static obstacles, such as mailboxes, bus stops, and fire hydrants.
3) Urban-Nav-3: scenes with dynamic obstacles, such as pedestrians and other mobile agents. Their trajectories are created using the proposed strategy for generating interactive dynamics.
Overall, we generate 768 (256$\times$3) interactive urban scenes for training (Urban-Nav-i-Train) and 768 (256$\times$3) scenes for testing (Urban-Nav-i-Test). Each scene is with the size of 10 $m$ $\times$ 10 $m$\footnote{\texttt{URBAN-SIM} can generate infinite scenes with any size. Here, we set a fixed scene number and size to standardize benchmarks.}.

\noindent
\underline{\textit{2) Urban-Loc.}} To benchmark the urban locomotion tasks, we generate 4 subsets of terrain conditions based on the \texttt{URBAN-SIM}.
1) Urban-Loc-1: flat ground.
2) Urban-Loc-2: slope terrains with different sampled dip angles and lengths.
3) Urban-Loc-3: stair terrains with different sampled stair numbers and heights.
4) Urban-Loc-4: rough terrains with different degrees of the ground's bumps and dips.
In Table~\ref{tab:urban_loc_para}, we give the parameter range of different subsets. In the test set, the parameter range is different from the training set to make out-of-distribution evaluations.
Overall, we generate a space with $30\times30$ regions for each type of terrain in the training stage, each of which has a different difficulty level, and train the locomotion using curriculum learning. During the testing stage, we increased the overall difficulty and spawned $16\times16$ regions, uniformly placing 256 agents to test the performance of the model. 

\begin{table}[h!]
\centering
\caption{\textbf{Parameter sampling of the terrain generation.}}
\label{tab:urban_loc_para}
\resizebox{\columnwidth}{!}{%
\begin{tabular}{c|c|c}
\toprule
\textbf{} & \textbf{Training} & \textbf{Testing} \\
\midrule
Urban-Loc-1 (Flat) & - & - \\
Urban-Loc-2 (Stair) & $\sim U(0.05, 0.23)$ & $\sim U(0.10, 0.30)$ \\
Urban-Loc-3 (Slope) & $\sim U(0.00, 0.40)$ & $\sim U(0.05, 0.80)$ \\
Urban-Loc-4 (Rough) & $\sim U(0.02, 0.10)$ & $\sim U(0.05, 0.20)$ \\
\bottomrule
\end{tabular}
}
\end{table}

\noindent
\underline{\textit{3) Urban-Tra-Standard.}}
To quantitatively benchmark the kilometer-scale urban traverse task, we build a complex running-track testing environment with an unseen combination of layouts, obstacles, terrains, and dynamics.
The running track is 120 $m$ in length and 10 $m$ in width. It is separated into 6 units, each unit with 20 $m$ length and 10 $m$ width.The 6 scenarios include two parts: even surfaces and uneven surfaces; each part covers three types of situations: unobstructed ground, static obstacles, and dynamic obstacles.
For each robot, we set a circuitous traversable route of the same length based on its traversability. Route length is set as 1,200 $m$.
Urban-Tra-Standard covers intricate conditions that will be encountered in kilometer-scale micromobility tasks in urban spaces. This scene supports an agile and all-sided evaluation of robots' performance in urban traverse tasks.

\noindent
\underline{\textit{4) Urban-Tra-City.}} To qualitatively evaluate robot behaviors on the urban traverse task, we further construct a city-scale testing environment with an area of 1,440,000 $m^2$ (1,200 $m$ width $\times$ 1,200 $m$ length).
This scene features vivid and realistic urban scenarios, including diverse layouts, complex terrains, city facilities, traffic flows, and movements of dense pedestrians. These elements present a challenging and intricate evaluation environment for micromobility devices in the urban traverse task.
For dynamics, environmental agents (\eg, pedestrians and mobile machines) are controlled by a multi-agent path plan algorithm ORCA~\cite{van2011reciprocal}, and vehicles are controlled by IDM policy~\cite{treiber2000congested}. Finally, we regulate all dynamic agents to make them comply with several traffic rules, such as traffic lights and speed limit signs.

\subsection{Evaluation Metrics}

For all three tasks, we evaluate from three aspects -- Completeness, Efficiency, and Safety.
For the urban locomotion task, we use 4 evaluation metrics.
Completeness: Balance ($\%$);
Efficiency: $x$-displacement (m) and Time to fall (s) - TTF;
Safety: Smoothness.
For the urban navigation task, we use 5 evaluation metrics.
Completeness: Success Rate ($\%$) and Route Completion ($\%$).
Efficiency: Success weighted by Path Length - SPL.
Safety: On Walkable Region (\%) and Collision Times.
For the urban traverse task, we design 5 metrics.
Completeness: Attempts to Success.
Efficiency: Labor Cost (s), Human Intervention Times, Mechanical Cost of Transport - MCoT, and Moving Speed (m/s).
Safety: Collision Times. We list the meaning and calculation formula of each metric in Table \ref{tab:urban_tasks_metrics}.

\begin{table*}[h!]
\centering
\caption{\textbf{Evaluation metrics for urban tasks.}}
\label{tab:urban_tasks_metrics}
\resizebox{\textwidth}{!}{%
\begin{tabular}{c|c|c}
\toprule
\textbf{Metric Name} & \textbf{Description} & \textbf{Calculation Formula} \\
\midrule
\multicolumn{3}{c}{\textbf{Urban Locomotion Task}} \\
\midrule
 Balance & \makecell{Whether stability is maintained \\ within a specified time} & $\frac{1}{N}\sum_{i=1}^{N}b_i$ \\
 \hline
$x$-displacement & \makecell{Displacement in the velocity direction under \\ fixed magnitude and random orientation}  & $E[({x_{fall}}-x_{start})\cdot cos(\theta_{target})]$  \\
 \hline
TTF & \makecell{Time interval from the start of \\ motion to loss of balance} &$E[t_{F}-t_{S}]$ \\
 \hline
Smoothness & \makecell{Smoothness of the motion sequence} & $E[\frac{1}{t_{F}-t_{S}}\sum_{t=2}^{t_{F}-t_{S}} ||torque_{t}-torque_{t-1}||]$ \\
\midrule
\multicolumn{3}{c}{\textbf{Urban Navigation Task}} \\
\midrule
Success Rate &\makecell{The ratio of episodes where\\ the
agent arrives at the destination} & $\frac{1}{N}\sum_{i=1}^{N}s_i$\\
 \hline
Route Completion & \makecell{Average of percentage of \\ route completed in episodes} & $E[\frac{l_{moving}}{L_{route}}]$ \\
 \hline
SPL & \makecell{Success weighted by Path Length,  measuring \\ the efficiency of the path taken by the agent }& $E[\frac{l_{moving}}{L_{best}^{*}}\cdot s]$ \\
 \hline
On Walkable Region & \makecell{Average of the ratio of time when \\ the agent runs on walkable regions} & $E[\frac{1}{T_F-T_S}\sum_{t=1}^{T_F-T_S}1_{walkable}(t)]$ \\
 \hline
Collision & \makecell{Average of the ratio of time when \\ the agent collides with any objects}  & $E[\frac{1}{T_F-T_S}\sum_{t=1}^{T_F-T_S}1_{collision}(t)]$ \\
\midrule
\multicolumn{3}{c}{\textbf{Urban Traverse Task}} \\
\midrule
Attempts to Success & \makecell{Number of times the agent \\ resets in the task} & $\sum_{t=1}^{T_F-T_S}1_{reset}(t)$  \\
 \hline
Labor Cost & \makecell{Total time spent by humans \\ intervening in the task} & $\sum_{t=1}^{T_F-T_S}1_{intervene}(t)\cdot dt_{interval}$\\
 \hline
Human Intervention Times & \makecell{Number of times 
 humans \\ intervene in the task} & $\sum_{t=2}^{T_F-T_S}1_{intervene}(t)\cdot(1-1_{intervene}(t-1))$  \\
 \hline
 MCoT& \makecell{Mechanical Cost of Transport\\Energy cost for moving per unit weight and distance} & $\frac{1}{T_F-T_S}\sum_{t=1}^{T_F-T_S} \sum_i \frac{max(torque_i\cdot v_i, 0)}{m\cdot g\cdot v_{body}}$ \\
 \hline
Speed & \makecell{Average speed in the task} &$\frac{1}{T_F-T_S}\sum_{t=1}^{T_F-T_S}||x_t-x_{t-1}||$ \\
 \hline
Collision Times & \makecell{Number of times the agent \\ collides with any objects in the task} & $\sum_{t=1}^{T_F-T_S}1_{collision}(t)$\\
\bottomrule
\end{tabular}
}
\end{table*}

\subsection{Models}

\paragraph{Urban locomotion.}
In urban locomotion, the embodied AI agent controls the robot's locomotion, ensuring stable and efficient movement across various terrains such as flat surfaces, slopes, and stairs.
The input is an elevation map, sensorimotor signals (such as joint positions, velocities, and torques), and velocity commands from the navigation module. The output is low-level control signals (such as joint positions and wheel velocity), which a controller then translates into actual motor commands that drive the robot's actuators.
We learn four locomotion models on different subsets -- Urban-Loc-i-Train (j=1,2,3,4), as well as a general locomotion model in the mixed terrains, which will be used in Human-AI-Mode-2 in the urban traverse benchmark.
For each model, the focus is to match the target velocity command from the navigation module while keeping balance on the terrain.
We model locomotion as a reinforcement learning problem, where the AI learns an optimal locomotion policy $\pi_{\text{loc}}^{j}(s) \in \Pi_{\text{loc}}$ through PPO~\cite{schulman2017proximal}.
The locomotion policy is trained to maximize rewards that prioritize smooth, stable movement:
\begin{equation}
    \pi_{\text{loc}}^{j*}(s) = \arg \max_{\pi_{\text{loc}}^j} \mathbb{E} \left[ \sum_{t=0}^{T} \gamma^t r_{\text{loc}}^j(s_t, a_t) \right],
\end{equation}
where $r_{\text{loc}}^j(s_t, a_t)$ is the reward function for locomotion, which provides rewards for maintaining balance, smooth movement, and energy efficiency, penalties are applied for unsafe actions (\eg, stumbling or wasting energy); $s_t$: The state of the robot’s body at time $t$; The locomotion action at time $t$ -- low-level control signals to drive robot's actuators; $\gamma$: controls the balance between immediate rewards and long-term rewards.
Finally, we can get a set of primitive locomotion policies, each of which excels in maintaining stability on one specific terrain when advancing toward the target point.

\paragraph{Urban navigation.}
In urban navigation, the embodied AI agent handles local navigation, determining how the robot should move to stay within traversable areas while avoiding obstacles and pedestrians.
The input is RGBD frames, the current coordinate of the agent, and the coordinate of the target point. The output is the target linear and angular velocity command.
We learn three navigation models on different subsets -- Urban-Nav-j-Train (j=1,2,3).
For each policy, the goal is to maximize both efficiency and safety while progressing toward the goal. This problem is modeled as a Markov Decision Process (MDP)~\cite{puterman1990markov}, where the AI learns to optimize its navigation policy $\pi_{\text{nav}}^{i}(s) \in \Pi_{\text{nav}}$ using Proximal Policy Optimization (PPO)~\cite{schulman2017proximal}, a reinforcement learning method.
The navigation policy is trained to maximize the expected cumulative rewards:
\begin{equation}
    \pi_{\text{nav}}^{i*}(s) = \arg \max_{\pi_{\text{nav}}^i} \mathbb{E} \left[ \sum_{t=0}^{T} \gamma^t r_{\text{nav}}^i(s_t, a_t) \right],
\end{equation}
where $r_{\text{nav}}^i(s_t, a_t)$: The reward function for navigation, providing positive rewards for progress towards the target and safe navigation (\eg, avoiding obstacles), while penalizing collisions or inefficiency; $s_t$: The state of the robot at time $t$; $a_t$: The navigation action at time $t$ -- the linear and angular velocity of robot. $\gamma$ controls the balance between immediate rewards and long-term rewards.
With the optimization of policies, we can finally get a set of primitive navigation policies, each of which is an expert in dealing with one specific navigation scenario.

\paragraph{Urban traverse.}
In urban traverse, humans and AI cooperate to complete the kilometer-scale task. Humans need to make a decision that involves determining whether the human or AI should control the robot based on the observed conditions (terrain and obstacle types) of the next interval. The interval can be split by geodesic distance or time duration.
The process is split into two main decisions: 1) whether the human supervisor should take over or let the AI continue, and 2) if the AI is in control, selecting the best combination of navigation and locomotion experts.
The decision-making process can be expressed as:
\begin{equation}
    \pi_{\text{system}}^*(s_k) = \begin{cases} 
        \pi_h(s_k) & \text{if } h_k = 1, \\
        \left( \pi_{\text{nav}}^{i^*}, \pi_{\text{loc}}^{j^*} \right) 
        & \text{if } h_k = 0,
    \end{cases}
\end{equation}
where $s_k$ is the current \textit{state} of the system at decision point $k$, $h_k$ is a binary variable representing whether human control is selected ($h_k = 1$) or AI control is maintained ($h_k = 0$); $\pi_h(s_k)$ represents the human control policy; $\pi_{\text{nav}}^{i^*}$ is the selected primitive navigation policy from the set $\Pi_{\text{nav}} = \{\pi_{\text{nav}}^1, \pi_{\text{nav}}^2, \dots, \pi_{\text{nav}}^{N_{\text{nav}}}\}$; $\pi_{\text{loc}}^{j^*}$ is the selected primitive locomotion policy from the set $\Pi_{\text{loc}} = \{\pi_{\text{loc}}^1, \pi_{\text{loc}}^2, \dots, \pi_{\text{loc}}^{N_{\text{loc}}}\}$.

For the urban traverse task, based on the foundational models, we construct 4 different control modes, spanning from fully human to fully AI as below.

\noindent
\underline{\textit{1) Human.}} A fully human control mode. Humans control a robot's navigation to make waypoint following. The robot uses a general locomotion model trained on mixed terrain.

\noindent
\underline{\textit{2) AI.}} A fully AI control model. The robot uses a general navigation model to make waypoint following and a general locomotion model to traverse terrains.

\noindent
\underline{\textit{4) Human-AI-Mode-1.}} A human AI shared control mode. Humans dispatch foundational navigation models and locomotion models based on scene conditions at every decision point or call humans for takeover.

\noindent
\underline{\textit{3) Human-AI-Mode-2.}} A human AI shared control mode. Humans dispatch foundational navigation models and the general locomotion model based on scene conditions at every decision point or call humans for takeover.

\subsection{Navigation Benchmark}
In this section, we will describe the experiment in detail regarding the task of navigation.

\paragraph{Environments.} There are three environments, i.e., NavClear, NavStatic and NavDynamic. For experiments on NavClear, there is nothing except the ego agent in the environment. The agent needs to move from the starting point to the destination, given the location of the ending point. We train the agent on $ 10 m\times10 m$ 
 regions with different textures of the walkable and non-walkable regions. To evaluate the trained policy, we generate $15 m\times15 m$ regions with the same textures as the training domain.

 \noindent For experiments on NavStatic, there are only static objects except the ego agent in the environment. The agent needs to move from the starting point to the destination, given the location of the ending point, and it should collide with any other objects as little as possible. We train the agent on $ 10 m\times10 m$ 
 regions with different textures of the walkable and non-walkable regions and $4$ objects. To evaluate the trained policy, we generate $ 15 m\times15 m$ regions with the same textures as the training domain and $8$ objects.

 \noindent For experiments on NavDynamic, there are not only static objects but also dynamic pedestrians, except the ego agent, in the environment. The agent needs to move from the starting point to the destination, given the location of the ending point, and it should collide with any other objects as little as possible. We train the agent on $10m \times 10 m$ 
 regions with different textures of the walkable and non-walkable regions, $4$ objects, and $2$ pedestrians. To evaluate the trained policy, we generate $15 m\times15 m$ regions with the same textures as the training domain, $8$ objects, and $3$ pedestrians.

\paragraph{Action spaces.} We use the action that is a 2-dimension vector normalized to $[-1.0, 1.0]$, where the two components indicate the relative proportions of the max velocity in the $x-$ and $y-$ directions. It would be processed by the locomotion model for Unitree robots or the Ackermann model for the COCO robot to generate the final action as input of the physics model.

\paragraph{Observation spaces.} Multi-model observations are used in all robots for training, including RGB and depth images with the resolution of $128\times 128$, the vector indicating the localization and the destination, the projected velocity on the target direction, and the height map in the region of $1.6 m\times1.0 m$.

\paragraph{Methods.} In our study, we employ the Proximal Policy Optimization (PPO) \cite{schulman2017proximal} to train the policy for all agents. PPO is a widely adopted and effective method, and it is easy to scale by
adopting parallel and distributed training. It took about 8, 18, and 24 hours to train the model on a single Nvidia L40S GPU for NavClear, NavStatic, and NavDynamic, respectively.  The detailed hyperparameters are provided in Table \ref{tab:ppo-hyper}.

\begin{table}[!h]
    \centering
    \caption{\textbf{Hyper-parameters of RL in urban navigation.}}
    \vspace{0.1in}
    \begin{tabular}{l l} \toprule  
    \textbf{PPO Hyper-parameters} & \textbf{Value} \\
    \hline 
    Environmental horizon $T$ &200\\
         Learning rate&4e-4\\
 Discount factor $\gamma$&0.99\\
 GAE parameter $\lambda$&0.95\\
  Clip parameter $\epsilon$ & 0.2\\
         Train batch size & $256 \times 60 (n_{env}\times n_{step})$ \\
 SGD minibatch size&3840\\
         Value loss coefficient&1.0\\ 
         Entropy loss coefficient&0.0005\\ \bottomrule
    \end{tabular}
    \label{tab:ppo-hyper}
\end{table}

\paragraph{Rewards.} 
The reward function is composed as follows:
\begin{equation}
    R = R_{term} + c_1 R_{track} + c_2 R_{walkable} + c_3 R_{collision}
\end{equation}
\begin{itemize}
\item Terminal reward $R_{term}$: 
 a sparse reward set to $+50$ if the vehicle reaches the destination, and $-100$ for out of boundaries. If given $R_{term}\neq 0$ at any time step $t$, the episode will be terminated at $t$ immediately. 

 \item Tracking reward $R_{track}$: a dense reward defined as $R_{track}=[1 - torch.tanh(d_{s-2-d} / 1.0)] + 0.2 * [1 - torch.tanh(d_{s-2-d} / 0.2)]$, wherein the $d_{s-2-d} $ denotes the distance between the location and the destination. We set the weight of $R_{track}$ as $c_1=2.0$.

  \item On-walkable-region reward $R_{walkable}$: a dense reward defined as $R_{walkable}=-1_{non-walkable}(t)$, wherein the $1_{non-walkable}(t)$ denotes that the agent walks on non-walkable regions at time $t$. We set the weight of $R_{walkable}$ as $c_2=0.5$.

  \item Collision reward $R_{collision}$: a dense negative reward defined as $-1(c_{t})$, wherein the $c_{t}$ denotes the collision between agents and any other objects at time $t$ and $1(\cdot)$ is the indicator function. We set the weight of $R_{crash}$ as $c_3=1.0$.
\end{itemize}

\subsection{Locomotion Benchmark}
\label{locomotion_benchmark_supp}
In this section, we will describe the experiment in detail regarding the task of locomotion.

\paragraph{Environments.} There are several environments, \ie, LocoFlat, LocoSlope, LocoStair, LocoRough, and general environment. Each environment has a specific type of terrain, and the general one has all types of terrain. We provide details of the environments for locomotion training in Table \ref{tab:urban_loc_para}.

\paragraph{Action spaces.} The action space is specifically given by the URDF file of the robot; there are 12, 16, and 37 joints for go2, g1, and b2w, respectively.

\paragraph{Observation spaces.} Multi-model observations are used in all robots for training, including linear velocity, angular velocity, target velocity, joint position, joint velocity, latest action, and the height map in the region of $1.6 m\times1.0 m$.

\paragraph{Methods.} Like the navigation benchmark, we employ the Proximal Policy Optimization (PPO) \cite{schulman2017proximal} to train the policy for all agents. PPO is a widely adopted and effective method, and it is easy to scale by
adopting parallel and distributed training. It took about 8, 18, and 24 hours to train the model on a single Nvidia L40S GPU for NavClear, NavStatic, and NavDynamic, respectively. The detailed hyperparameters are provided in Table \ref{tab:ppo-hyper}.

\paragraph{Rewards.} 
The reward function is composed as follows:
\begin{equation}
    R = R_{term} + c_1 R_{track} + c_2 R_{vel} + c_3 R_{smooth}
\end{equation}
\begin{itemize}
\item Terminal reward $R_{term}$: a sparse reward set to $-100$ is needed for situations where there is a contact force $> 1.0$ between the agent and the ground. If given $R_{term}\neq 0$ at any time step $t$, the episode will be terminated at $t$ immediately. 

 \item Tracking reward $R_{track}$: a dense reward defined as $R_{track}=\exp(-\delta v / 0.25**2)+0.5\times \exp(-\delta z / 0.25**2)$, wherein the $\delta v $ denotes the error between the current velocity and target velocity. We set the weight of $R_{track}$ as $c_1=1.0$.

  \item On-walkable-region reward $R_{vel}$: a dense reward defined as $R_{vel}=-v_z^2$, wherein the $v_z$ denotes along the $z-$axis. We set the weight of $R_{walkable}$ as $c_2=2.0$.

  \item Collision reward $R_{smooth}$: a dense negative reward defined as $-||a(t)-a(t-1)||_2^2$, wherein the $a(t)$ denotes the action of the agent at time $t$ . We set the weight of $R_{crash}$ as $c_3=0.01$.
\end{itemize}

\begin{table}[!h]
    \centering
    \caption{\textbf{Hyper-parameters of RL in urban locomotion.}}
    \vspace{0.1in}
    \begin{tabular}{l l} \toprule  
    \textbf{PPO Hyper-parameters} & \textbf{Value} \\
    \hline 
    Environmental horizon $T$ &1000\\
         Learning rate&1e-3\\
 Discount factor $\gamma$&0.99\\
 GAE parameter $\lambda$&0.95\\
  Clip parameter $\epsilon$ & 0.2\\
         Train batch size & $1024 \times 24 (n_{env}\times n_{step})$ \\
 SGD minibatch size&24576\\
         Value loss coefficient&1.0\\ 
         Entropy loss coefficient&0.0005\\ \bottomrule
    \end{tabular}
    \label{tab:ppo-hyper}
\end{table}

\subsection{Urban Traverse Benchmark}

Table~\ref{tab:urban_traverse_benchmark} brings the following insights:
\textit{1) Human mode ensures best safety and completeness, but at high labor costs.} The fully human control mode achieves the fewest Trials to Success (4) and lowest Collision Times (15), showcasing its reliability in safety and task completion. However, it incurs the highest Labor Cost (1189.27 s) and Human Intervention Times (14), highlighting the scalability challenges of full human control.
\textit{2) AI mode maximizes efficiency but struggles with safety.} The fully AI mode eliminates human intervention and achieves the lowest Labor Cost (0 s) but performs poorly in safety, with the highest Collision Times (49) and Trials to Success (66). This highlights the current limitations of AI in complex, dynamic urban environments.
\textit{3) Shared control balances efficiency and safety.} Shared control modes, especially Human-AI-Mode-1, strike a balance between performance metrics, with reduced labor costs (189.52 s) and moderate intervention times (67). Human-AI-Mode-2 achieves the highest Speed (1.01 m/s), indicating enhanced efficiency. These modes significantly reduce human effort compared to the human mode while maintaining better safety than the fully AI mode.

\begin{table}[h!]
\small
\caption{\textbf{Urban traverse benchmark.} Different colors indicate the best performance of different aspects among four control modes: \legendsquare{colorbestLA} Completeness; \legendsquare{colorbestLB} Efficiency; \legendsquare{colorbestLC} Safety.}
\label{tab:urban_traverse_benchmark}
\begin{center}
\resizebox{0.47\textwidth}{!}{
    \begin{tabular}{c|cccc}
        \toprule
        \textbf{Metrics}  & \textbf{Human} & \textbf{Human-AI-Mode-1} & \textbf{Human-AI-Mode-2} & \textbf{AI} \\
        \midrule
        Trials to Success $\downarrow$ & \colorbestLA{4} & 27 & 58 & 66 \\
        \midrule
        Labor Cost (s) $\downarrow$ & 1189.27 & 189.52 & 187.52 & \colorbestLB{0.00} \\
        Human Inter. Times $\downarrow$ & 14 & 67 & 32 & \colorbestLB{0} \\
        MCoT $\downarrow$ & \colorbestLB{0.65} & 0.76 & 1.03 & 0.99 \\
        Speed (m/s) $\uparrow$ & 0.76 & 0.96 & \colorbestLB{1.01} & 0.98 \\
        \midrule
        Collision Times $\downarrow$ & \colorbestLC{15} & 20 & 32 & 49 \\
        \bottomrule
    \end{tabular}
}
\end{center}
\end{table}

\subsection{Necessity of Foundational Tasks.}

\begin{figure}[h!]
    \centering
    \includegraphics[width=1\linewidth]{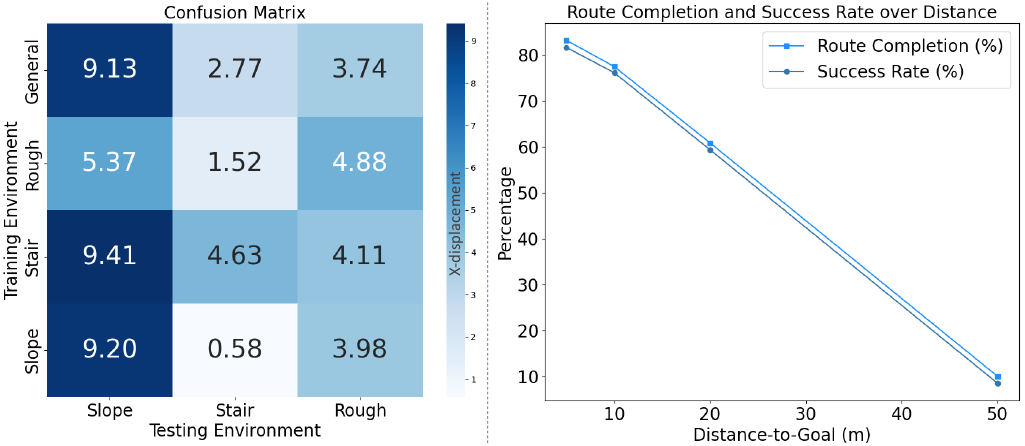}
    \vspace{-0.2in}
    \caption{\textbf{Necessity of foundational tasks.}
    }
    \vspace{-0.2in}
    \label{fig:evaluation_foundational_tasks}
\end{figure}

\textit{Settings.} To examine the importance of factorizing autonomous micromobility into essential tasks, we evaluate two aspects:
2) Task Conditions: We perform cross-evaluations on locomotion models trained under different ground conditions -- ``Slope'', ``Stair'', and ``Rough''. Further, we compare this to a general model trained on mixed conditions.
1) Task Size: We train navigation models with varying sizes of the scene, from 5$m$ $\times$ 5$m$ to 50$m$ $\times$ 50$m$, to analyze changes in learning difficulty.
\textit{Results.} 
1) Figure~\ref{fig:evaluation_foundational_tasks} (Left) demonstrates that domain-specific models trained under defined conditions (diagonal entries) outperform mixed-condition models. General models (4th row) struggle in specific settings except on``Slope'', suggesting that the increased complexity of terrains (``Stair'' and ``Rough'') makes mixed training less effective.
Separating the holistic task by conditions can simplify the learning of each subtask and make it easy to evaluate different aspects of model capabilities.
2) As shown in Figure~\ref{fig:evaluation_foundational_tasks} (Right), performance declines linearly as the training area size increases, indicating that learning a policy across large areas at one stock is highly challenging. A practical approach for long-horizon tasks could be linking smaller, foundational task models, as the proposed human-AI shared autonomous approach for the urban traverse task.

\section{Datasheet}
\label{sec:data_sheet}

This datasheet in Table~\ref{tab:data_sheet} provides a structured summary of the dataset associated with \texttt{URBAN-SIM}, designed to ensure transparency, reproducibility, and responsible use. Inspired by guidelines~\cite{Gebru2021DatasheetsFD}, it covers the dataset’s:
\textit{1) Motivation.} The dataset’s purpose and creators.
\textit{2) Composition.} Format, structure, and relationships of dataset instances.
\textit{3) Collection.} How data was generated and verified.
\textit{4) Uses.} Recommended applications and potential restrictions.
\textit{5) Distribution and Maintenance.} Plans for sharing, updates, and community contributions.

This documentation equips researchers with essential insights to use the dataset effectively and responsibly to advance autonomous micromobility research.

\onecolumn

\begin{longtable}{p{0.35\linewidth} |p{0.6\linewidth} }
    \bottomrule
        \multicolumn{2}{c}{\rule{0in}{0.2in}\textbf{Motivation}\vspace{0.10in}}\\
    \toprule
     For what purpose was the dataset created? & The dataset was created to enable large-scale robot learning on urban scenes and facilitate future autonomous micromobility research.\\[0.15in]
     \midrule
     Who created and funded the dataset? & This work was created and funded by the \texttt{URBAN-X} team.\\
    \bottomrule
        \multicolumn{2}{c}{\rule{0in}{0.2in}\textbf{Composition}\vspace{0.10in}}\\
    \toprule
    What do the instances that comprise the dataset represent? & Each instance is a JSON file, including the configuration of the scenes in \texttt{URBAN-SIM} and a specific seed.
    \\[0.15in]
    \midrule
    How many instances are there in total (of each type, if appropriate)? & There are 1,536 scenes released in the dataset derived from \texttt{URBAN-SIM}, along with the code to sample substantially more. \\[0.15in]
    \midrule
    Does the dataset contain all possible instances or is it a sample (not necessarily random) of instances from a larger set? & We offer 1,536 urban scenes and code of the Hierarchical Urban Generation pipeline that can create an infinite number of scenes.\\[0.15in]
    \midrule
    What data does each instance consist of? & Each instance is specified as a JSON file.\\[0.15in]
    \midrule
    Is there a label or target associated with each instance? & No.\\[0.15in]
    \midrule
    Is any information missing from individual instances? & No.\\[0.15in]
    \midrule
    Are relationships between individual instances made explicit (e.g., users' movie ratings, social network links)? & Each urban scene is created independently, so there are no connections between the scenes.\\[0.15in]
    \midrule
    Are there recommended data splits? & Yes. See Section~\ref{sec:bench_data} in the Appendix.\\[0.15in]
    \midrule
    Are there any errors, sources of noise, or redundancies in the dataset? & No.\\[0.15in]
    \midrule
    Is the dataset self-contained, or does it link to or otherwise rely on external resources (e.g., websites, tweets, other datasets)? & The dataset is self-contained.\\[0.15in]
    \midrule
    Does the dataset contain data that might be considered confidential? & No.\\[0.15in]
    \midrule
    Does the dataset contain data that, if viewed directly, might be offensive, insulting, threatening, or might otherwise cause anxiety? & No.\\
    \bottomrule
        \multicolumn{2}{c}{\rule{0in}{0.2in}\textbf{Collection Process}\vspace{0.10in}}\\
    \toprule
    How was the data associated with each instance acquired? & Each instance was created with the same urban generation pipeline, with different randomly sampled variables.\\[0.15in]
    \midrule
    If the dataset is a sample from a larger set, what was the sampling strategy? & The dataset consists of 1,536 scenes, each by sampling the difficulty parameters (such as dip angles of slopes and density of pedestrians).\\[0.15in]
    \midrule
    Who was involved in the data collection process? & The authors were the sole individuals responsible for creating the dataset.\\[0.15in]
    \midrule
    Over what timeframe was the data collected? & Data was collected in Oct. 2024.\\[0.15in]
    \midrule
    Were any ethical review processes conducted? & No.\\
    \bottomrule
        \multicolumn{2}{c}{\rule{0in}{0.2in}\textbf{Preprocessing/Cleaning/Labeling}\vspace{0.10in}}\\
    \toprule
    Was any preprocessing/cleaning/labeling of the data done? & For the 3D assets that are used on the scene generation, we will first manually check the visual quality, and scale before use.\\[0.15in]
    \midrule
    Was the ``raw'' data saved in addition to the preprocessed/cleaned/labeled data? & There is no raw data.\\[0.15in]
    \midrule
    Is the software that was used to preprocess/clean/label the data available? & All of the code related to preprocessing, cleaning, and labeling the data will be made available.\\
    \bottomrule
        \multicolumn{2}{c}{\rule{0in}{0.2in}\textbf{Uses}\vspace{0.10in}}\\
    \toprule
    Has the dataset been used for any tasks already? & Yes. See Section~\ref{sec:bench_data} of the Appendix.\\[0.15in]
    \midrule
    What (other) tasks could the dataset be used for? & The scenes can be used in a wide variety of tasks, such as autonomous micromobility, embodied AI, vision language models, computer vision, and urban accessibility.
    \\[0.15in]
    \midrule
    Is there anything about the composition of the dataset or the way it was collected and preprocessed/cleaned/labeled that might impact future uses? & No.\\[0.15in]
    \midrule
    Are there tasks for which the dataset should not be used? & Our dataset can be used for both commercial and non-commercial purposes.\\
    \bottomrule
        \multicolumn{2}{c}{\rule{0in}{0.2in}\textbf{Distribution}\vspace{0.10in}}\\
    \toprule
    Will the dataset be distributed to third parties outside of the entity on behalf of which the dataset was created? & Yes. We plan to make the entirety of the work open-source,  including the code used to create scenes, generate terrains and dynamics, and train agents. We will also release the related asset repositories.\\[0.15in]
    \midrule
    How will the dataset be distributed? & The scene files will be distributed using a custom Python package.\newline
    The code will be distributed on GitHub. The asset repositories will be distributed on Google Drive.\\[0.15in]
    \midrule
    Will the dataset be distributed under a copyright or other intellectual property (IP) license, and/or under applicable terms of use (ToU)? & The scene dataset, 3D asset repository, and code will be released under the Apache 2.0 license. \\[0.15in]
    \midrule
    Have any third parties imposed IP-based or other restrictions on the data associated with the instances? & Yes. 3D object assets: OmniObject3D~\cite{wu2023omniobject3d} is under CC BY 4.0 license. 3D human assets: Synbody~\citep{yang2023synbody} is under CC BY-NC-SA 4.0 license. \\[0.15in]
    \midrule
    Do any export controls or other regulatory restrictions apply to the dataset or to individual instances? & No.\\
    \bottomrule
        \multicolumn{2}{c}{\rule{0in}{0.2in}\textbf{Maintenance}\vspace{0.10in}}\\
    \toprule
    Who will be supporting/hosting/maintaining the dataset? & The authors will be providing support, hosting, and maintaining the dataset.\\[0.15in]
    \midrule
    How can the owner/curator/manager of the dataset be contacted? & For inquiries, email urban\_x\_team@gmail.com>.\\[0.15in]
    \midrule
    Is there an erratum? & We will use GitHub issues and Slack groups to track issues with the dataset.\\[0.15in]
    \midrule
    Will the dataset be updated? & We will continue adding support for new features to make the simulated urban scenes more diverse and realistic. We also intend to define and support new task training and evaluation in the future.\\[0.15in]
    \midrule
    If the dataset relates to people, are there applicable limits on the retention of the data associated with the instances (e.g., were the individuals in question told that their data would be retained for a fixed period of time and then deleted)? & The dataset does not relate to people. \\[0.15in]
    \midrule
    Will older versions of the dataset continue to be supported/hosted/maintained? & Yes. Revision history will be available for older versions of the dataset.\\[0.15in]
    \midrule
    If others want to extend/augment/build on/contribute to the dataset, is there a mechanism for them to do so? & Yes. The work will be open-sourced, and we intend to offer support to assist others in using and building upon the dataset.\\[0.15in]
\bottomrule
    \caption{A datasheet \citep{Gebru2021DatasheetsFD} for \texttt{URBAN-SIM} and its derived datasets.}
    \label{tab:data_sheet}
\end{longtable}

\twocolumn

\clearpage
\section{Discussion}
\label{sec:discussion}

\paragraph{Impact.}

This work on scalable urban simulation, combining \texttt{URBAN-SIM} and \texttt{URBAN-BENCH}, holds transformative potential across multiple domains, including Embodied AI, Urban Development, and Society.  

\noindent
1) \textit{Embodied AI.} The proposed platform and benchmarks push forward research in autonomous micromobility by addressing key challenges such as urban navigation, locomotion, and long-horizon planning. By providing realistic, scalable, and diverse environments, our work could accelerate advancements in robust AI systems capable of adapting to dynamic, complex urban spaces. This contributes to foundational areas like multi-agent systems, reinforcement learning, and embodied cognition.

\noindent
2) \textit{Urban development.} The simulation tools introduced in this work could revolutionize urban planning and infrastructure design. By simulating various traffic patterns, pedestrian flows, and robot interactions, urban spaces could be optimized for accessibility, safety, and efficiency. Applications in industries like last-mile delivery, public transportation, and assistive robotics could further drive innovation and enhance the operational effectiveness of urban services.

\noindent
3) \textit{Society.} This work fosters the safe and inclusive integration of autonomous robots into public spaces, supporting accessibility for mobility-impaired individuals and enabling technologies such as assistive wheelchairs and parcel delivery robots. By enhancing the capabilities of robots to navigate urban environments, this research could improve the quality of life and strengthen public services, contributing to more livable and equitable cities.

\noindent
4) \textit{Potential negative impacts.} The deployment of AI and robots in urban spaces, while promising, introduces risks. Increased automation may lead to job displacement and economic inequality, while privacy concerns arise with the presence of robots and sensors in public areas. Moreover, reliance on AI-driven systems raises challenges related to failure resilience and unintended societal consequences, such as altering human interactions in public spaces. Finally, the environmental costs of manufacturing and deploying robots and simulation technologies must be mitigated to ensure sustainable adoption. Addressing these issues is critical to ensuring the equitable and responsible advancement of autonomous micromobility.

\paragraph{Sim-to-real.}

\texttt{URBAN-SIM} is positioned as an Embodied AI simulator designed to enable rapid model training and evaluation before real-world deployment of physical robots. Following the standards set by popular embodied AI simulators, such as AI2-THOR~\citep{kolve2017ai2}, Habitat~\citep{savva2019habitat}, and ProcTHOR~\citep{deitke2022️procthor}, the current version focuses solely on simulation and does not include real-world experimentation. However, bridging the sim-to-real gap remains a key goal in our development roadmap.

To ensure \texttt{URBAN-SIM} evolves into a sustainable and impactful platform for the community, we are actively developing an end-to-end experimentation pipeline that extends simulation capabilities to real-world deployment.
Preliminary experiments with Unitree’s Go2 quadruped robot and COCO Robotics’ wheeled robot show promising results: training robots with abstract observations, such as depth maps, combined with domain randomization, has already achieved good transferability to real-world environments.

Looking ahead, we aim to provide comprehensive support for the ROS 2 interface, enabling seamless integration between \texttt{URBAN-SIM} and robots' control signals and sensor data streams.
Additionally, we plan to incorporate advances in neural rendering technologies, such as NeRF and Gaussian Splatting, to create highly realistic simulation environments, further reducing the sim-to-real gap and enhancing the applicability of trained models in real-world scenarios.

\paragraph{Limitations.} Although this work introduces a scalable framework for autonomous micromobility with benchmarks that address essential skills, two known limitations in its current design present opportunities for future work.

\noindent
1) \textit{Extension of robot capability.}
Our benchmarks primarily evaluate urban navigation, locomotion, and traverse, which are three essential skills for urban micromobility. However, additional capabilities, such as manipulation and multi-modal perception, are critical for enabling robots to perform more complex tasks, such as parcel delivery, trash collection, and human assistance.
Extending \texttt{URBAN-SIM} to support these capabilities would unlock new research directions and facilitate the development of mobile machines capable of offering more sophisticated services in urban environments.

\noindent
2) \textit{Real-world data distribution.}
While the procedurally generated scenes in \texttt{URBAN-SIM} offer extensive diversity, they do not fully reflect real-world data distributions. Extending \texttt{URBAN-SIM} to incorporate data from real-world sources, such as OpenStreetMap, could significantly enhance the realism of scene layouts and object placement. This integration would allow the platform to generate digital twins of actual cities, enabling more accurate and context-specific agent training.

\paragraph{Future work.}
We envision three primary directions for future work to further enhance and expand the capabilities of this work.

\noindent  
1) \textit{Building an open-source ecosystem.}
To foster collaboration and maximize the impact of our platform, we plan to actively develop an open-source ecosystem. We will begin by conducting surveys to understand user demands and refine our development roadmap accordingly. The platform’s code will be hosted on GitHub to encourage contributions from the broader research community. Additionally, we will establish communication channels, such as mailing lists and Slack groups, to gather real-time feedback and facilitate community engagement. To further promote the platform, we intend to organize workshops to showcase its capabilities and broaden its user base.

\noindent  
2) \textit{Enabling open-world learning.}
Our platform aims to support cutting-edge research in open-world learning, leveraging Vision-Language Models (VLMs) and Large-Language Models (LLMs) as heterogeneous agents in virtual urban environments. A promising direction is to endow agents with personal traits, such as jobs, personalities, and goals, using advances in LLMs~\cite{achiam2023gpt} and LVMs~\cite{liu2023visual}. These agents could spontaneously exhibit social~\cite{puig2023nopa} and interactive~\cite{park2023generative} behaviors, paving the way for more realistic and complex simulations of urban dynamics.

\noindent 
3) \textit{Addressing Current Limitations.}
We will continue to address the challenges identified in the sim-to-real gap and other limitations, such as extending the platform to incorporate additional robot capabilities and real-world data distributions.

In summary, our scalable urban simulation framework presents exciting new research opportunities. We are committed to the long-term development of \texttt{URBAN-SIM} and \texttt{URBAN-BENCH}, ensuring they serve as sustainable and impactful tools for advancing the field of embodied AI and autonomous micromobility.

\end{document}